\documentclass[10pt,twocolumn,letterpaper]{article}

\usepackage[algorithms]{wacv}   %

\usepackage{graphicx}
\usepackage{capt-of}
\usepackage{amsmath}
\usepackage{amssymb}
\usepackage{booktabs}
\usepackage{microtype}
\usepackage{multirow}
\usepackage{siunitx}
\DeclareSIUnit{\degree}{\ensuremath{{}^{\circ}}}
\DeclareSIUnit{\pixel}{px}
\DeclareSIUnit{\frame}{fr}
\usepackage{pgfplots}
\pgfplotsset{compat=1.17}

\definecolor{wacvblue}{rgb}{0.21,0.49,0.74}
\usepackage[pagebackref,breaklinks,colorlinks,allcolors=wacvblue]{hyperref}

\def\wacvPaperID{1120}
\def\confName{WACV}
\def\confYear{2027}

\newcommand{\ieeecopyrightyear}{2027}
\newcommand{\ieeenotice}{\copyright~\ieeecopyrightyear{} IEEE. Personal use of
  this material is permitted. Permission from IEEE must be obtained for all
  other uses, in any current or future media, including reprinting/republishing
  this material for advertising or promotional purposes, creating new
  collective works, for resale or redistribution to servers or lists, or reuse
  of any copyrighted component of this work in other works.}

\newcommand{\bigO}{\mathcal{O}}

\title{PadLoc: Training-Free Floor-Plan Localization by Sparse Geometric Matching}

\author{Marian Bookhahn \qquad Frank Neumann\\
        HTW Berlin -- University of Applied Sciences\\
        {\tt\small \{bookhahn, frank.neumann\}@htw-berlin.de}
}

\begin{document}
\AddToShipoutPicture*{\AtTextLowerLeft{\put(0,\LenToUnit{-0.3in}){%
  \parbox[t]{\textwidth}{\scriptsize\centering\ieeenotice}}}}
\twocolumn[{%
\renewcommand\twocolumn[1][]{#1}%
\maketitle
\begin{center}
  \centering
  \includegraphics[width=0.88\textwidth]{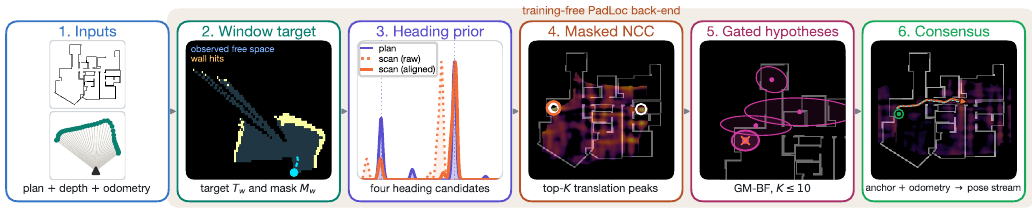}
  \captionof{figure}{\textbf{PadLoc method overview on real data} (tiles:
  intermediates of one trajectory chunk). A known floor plan, depth profiles,
  and planar odometry enter (1); the training-free back-end fuses a short
  window into a target and observation mask, which bounds drift (2), reduces rotation to four
  wall-angle heading candidates (3), scores translations with masked NCC and
  keeps top-$K$ peaks (4), propagates $K{\le}10$ gated Gaussian hypotheses
  with odometry (5), and re-reads the raw NCC fields by consensus to choose
  the global anchor and recover poses (6; dashed: ground truth).}
  \label{fig:teaser}
\end{center}
}]

\begin{abstract}
We propose \textbf{PadLoc}, a training-free localization back-end that places
a moving sensor within a known floor plan. Floor plans are cheap, persistent, and
robust to appearance change, but recent learned-depth methods localize against
them by filtering per-frame depth through dense $(x,y,\theta)$ pose grids.
PadLoc keeps the swappable depth front-end and replaces the dense grid back-end
with sparse geometric matching: dominant wall orientations define a small
pose-candidate set, masked normalized cross-correlation scores observed free
space against the plan, and a moving-pose Gaussian-mixture Bayes filter plus
consensus readout decodes the trajectory. Holding each competitor's depth fixed,
PadLoc matches or beats UnLoc's
forward filter and the F3Loc filter at every Gibson trajectory length, and
UnLoc's offline refinement from $T{=}35$ on; it exceeds Z-FLoc's published
results with a shared MapAnything front-end. On Bonn
LiLocBench, PadLoc improves pooled SR@1\,m over UnLoc by $23$--$32$\,pp under wheel
odometry and $35$--$43$\,pp with ground-truth motion; with full-FOV
LiDAR plus wheel odometry the same readout reaches
$\SI{100}{\percent}$ SR@1\,m at \SIrange{9}{10}{\centi\meter} median error on
the static sequence. On the Hilti-Trimble Localization Challenge PadLoc
localizes a camera-only fisheye rig on construction floors hundreds of metres
across at $0.90$--$0.98$ online SR@2\,m on three of five sessions; adding
learned wall isolation, applied to every session, covers all five at $0.67$--$0.92$
with the localizer unchanged throughout.
PadLoc thus matches or outperforms dense grid back-ends
while preserving sensor modularity and a compact state that does not grow with
trajectory length.
\end{abstract}

\section{Introduction}
\label{sec:intro}

Localizing a moving sensor against a 2D floor plan is attractive: floor
plans already exist, are compact, and avoid dense 3D mapping. Recent systems
such as F3Loc~\cite{chen2024f3loc} and UnLoc~\cite{unloc2026} pair learned
per-column depth with a dense Bayes-filter recursion over an $(x,y,\theta)$
grid---robust, but the posterior is large and persistent, and the design
is tied to one depth front-end.

PadLoc keeps the swappable depth front-end and changes only the localizer.
The front-end may be learned; the localizer itself has no trained parameters.
Dominant wall orientations form a small heading set, weighted masked
NCC~\cite{padfield2012masked} scores each sliding window, and the whole
NCC field over translations is kept, not only its argmax. The
top-$K$ local maxima feed a moving-pose Gaussian-mixture Bayes filter (GM-BF),
and a consensus readout reuses the raw NCC fields to select the absolute
trajectory anchor.

This separation lets us hold each competitor's depth fixed and compare only the
localizer. On Gibson, PadLoc matches or outperforms the UnLoc and F3Loc forward
filters at every trajectory length under a frozen readout, and UnLoc's offline
refinement from $T{=}35$ on; it exceeds Z-FLoc's published
results~\cite{zfloc2026}, with the largest gains at short horizons. The same matcher transfers unchanged
to Bonn LiLocBench with LiDAR, wheel odometry, VO, and RGB-D front-ends,
beating UnLoc on shared monocular depth, and scales to construction-size
buildings on the Hilti-Trimble Localization Challenge from a camera-only
fisheye front-end once the fusion window is matched to the reconstruction
batch.

We make three contributions.
\begin{itemize}
  \item \textbf{A training-free masked-NCC matcher.} Floor-plan localization
  cast as registration with partial overlap: an interior mask and an observation
  mask separate observed free space from unexplored cells, and rotation is
  reduced to dominant wall-orientation candidates.
  \item \textbf{A sparse multi-hypothesis decoder, online at bounded lag.}
  Top-$K$ NCC peaks become a compact posterior decoded by a moving-pose GM-BF
  and a consensus readout kept as a running sum. A single fixed lag bounds
  \emph{both} heading and translation evidence; the mixture is $\bigO(K)$, the
  consensus state does not grow with the trajectory, and the
  $\bigO(N\log N)$ matching kernel scales to building-sized plans and runs
  on a phone.
  \item \textbf{Matched-depth and cross-building evidence.} Holding each
  competitor's depth fixed, PadLoc matches or improves over the F3Loc and
  UnLoc forward filters and Z-FLoc's published results, and
  transfers across buildings. Knobs that depend on the
  front-end (fusion window, motion noise, cell size, blur)
  are set by stated rules, never per scene, and then carried from apartment scale (Bonn) to
  construction scale (Hilti) with a new MapAnything fisheye front-end.
\end{itemize}

\section{Related Work}
\label{sec:related}

\paragraph{Floor-plan localization.}
F3Loc~\cite{chen2024f3loc} and UnLoc~\cite{unloc2026} are the closest
learned-depth baselines on Gibson. Both pair learned per-column depth
predictions with a classical dense recursive Bayes filter over
$(x,y,\theta)$; UnLoc adds Depth-Anything-v2~\cite{yang2024depthanythingv2}
depth, aleatoric ray uncertainty, and offline SE(2) refinement. Earlier work includes learned
layout retrieval~\cite{lalaloc,lalalocpp}, latent floor-plan
matching~\cite{laser}, monocular layout filtering~\cite{boniardi2019}, and
RGB-D floor-plan localization~\cite{winterhalter}. Z-FLoc with
MapAnything~\cite{keetha2026mapanything} uses completed multiview depth and
batched geometric matching~\cite{zfloc2026}, so
we compare to it only in the batched regime. PadLoc instead keeps a sparse
translation posterior, replaces dense orientation filtering with dominant
wall-orientation candidates, and evaluates the same back-end under matched
learned-depth front-ends.

\paragraph{Classical multi-hypothesis localization.}
Global localization in a known map is classically solved by Monte Carlo
localization~\cite{dellaert1999mcl,fox2003kld,thrun2005probabilistic}---a
sampled pose posterior---or by multi-hypothesis Gaussian trackers with
MHT-style management~\cite{jensfelt2001active} built on Gaussian-sum
filtering~\cite{alspach1972gaussian}. Our matcher is correlative in the
sense of Konolige and Chou~\cite{konolige1999markov} and
Olson~\cite{olson2009realtime}: rotation in an outer loop, exhaustive
translation scoring of a rasterized reference, a score surface retained
beyond its argmax. Correlative and branch-and-bound
matching~\cite{olson2009realtime,hess2016cartographer} register a scan
within a prior-centred window and compress the surface to one Gaussian;
PadLoc searches the entire plan at $\bigO(N\log N)$ from a uniform prior,
collapses rotation to dominant wall-orientation candidates, normalizes
scores over observed cells only, and keeps the surface
as top-$K$ modes in a gated Gaussian mixture.

\paragraph{Masked correlation and orientation priors.}
Padfield~\cite{padfield2012masked} introduced Fourier-domain masked NCC. We
adapt it to weighted masks (\S\ref{sec:method_ncc}), with a building-interior
mask and an observed-support mask: observed
free space penalizes overlapping reference walls, while unobserved cells drop
out. The contribution is to expose the full translation response as a
multi-hypothesis posterior for temporal decoding. Similar sparse
representations appear in mixture filters and factor-graph
smoothers~\cite{olson2013inference,kaess2012isam2}, but our observations are
local maxima of a masked correlation surface. The orientation prior is
classical as well: Manhattan-world and dominant-direction assumptions are
standard in indoor layout estimation~\cite{coughlan1999manhattan} and scan
matching~\cite{konolige1999markov,olson2009realtime}. PadLoc uses that prior
to remove the dense $\theta$ axis while retaining multiple directed
candidates.

\paragraph{Swappable front-ends.}
PadLoc consumes per-frame depth profiles and relative planar odometry from
available front-ends---wheel encoders, LiDAR scan matching, or visual-inertial
odometry robust to dynamic scenes~\cite{dynamicvins}---rather than training a
sensor stack: ray-cast depth, learned F3Loc or UnLoc depth, 2D LiDAR, RGB-D,
and a \emph{batched} reconstruction front-end (MapAnything fisheye depth on
Hilti~\cite{helmberger2022hilti}), which only requires matching the fusion
window to the reconstruction batch (\S\ref{sec:exp_hilti}). Published Hilti
floor-plan entries optimize the full sequence offline---the winning entry
anchors an OKVIS2-X pose graph with Z-FLoc priors via whole-sequence
PGO~\cite{umemura2026floorplananchored}---whereas PadLoc emits poses online
at bounded lag.

\section{Method}
\label{sec:method}

\subsection{Overview}
\label{sec:method_overview}
PadLoc localizes a moving sensor in a known 2D floor plan from a
per-frame depth profile $d_t\in\mathbb{R}^{N_{\mathrm{col}}}$ and
relative planar odometry $\Delta_{t\rightarrow t+1}$.

Let $R_{\mathrm{raw}}$ be the rasterized wall map at resolution $\delta$,
$M_R$ its binary in-building mask, and $R=M_R\odot R_{\mathrm{raw}}$ the
premasked reference. Odometry gives cumulative planar motion
$c_t=(x_t,y_t)$ in a local gauge --- the odometry frame is known only up to
an unobservable global rigid transform that aligns it to the map. PadLoc
estimates a small candidate set
$\Theta=\{\theta_b\}_{b=1}^{B}$, builds a target raster
$T_w^b$ and validity mask $M_w^b$ for each sliding window $w$ and
heading $b$, and computes a masked NCC field $F_w^b(p)$ over translation
cells $p=(r,c)$. A field peak at $p$ places window $w$'s centroid at
map cell $p$ under heading $\theta_b$.

\subsection{Rasterization and Headings}
\label{sec:method_raster}
The floor plan is rasterized once: vector walls are drawn with
Bresenham rasterization, while pre-rasterized maps supplied at a
finer cell size are max-pooled down to the matching resolution. We fill a concave hull of the
wall pixels to obtain $M_R$ and use a blurred wall distance field
only to give reference wall strokes width comparable to depth noise
(Fig.~\ref{fig:teaser}, panel~1; supplement).

Each depth column is projected along its bearing into a local scan. A wall hit
sets an occupied endpoint ($T_w^b{=}1$, $M_w^b{=}w_{\mathrm{wall}}$), traversed
cells become observed free ($T_w^b{=}0$, $M_w^b{=}1$), and unreached cells stay
unobserved ($M_w^b{=}0$). Thus observed-free zeros penalize reference walls,
whereas unobserved zeros drop out. Wall hits use
$w_{\mathrm{wall}}{=}55$ so sparse wall returns dominate the masked NCC
statistics. If the depth front-end exposes a per-column aleatoric scale
$\hat b$, columns with $\hat b>\SI{0.30}{\meter}$ are removed before
rasterization. Both constants are fixed once on Gibson and never re-tuned; the supplement lists
all parameters and shows broad sensitivity plateaus, and the drop is ablated
in \S\ref{sec:exp_ablation}.

For heading candidates, PadLoc runs a causal pre-pass: every $5$ frames it
accumulates a $5$-frame scan, fits Hough wall segments, and votes segment
lengths into a wall-angle histogram modulo $\pi$. Cross-correlating target and
reference histograms yields gauge rotations; a lightweight tracker (a
von-Mises mixture smoothed online by an exponential moving average, EMA,
with forgetting) follows them, and
mirror expansion adds each $180^\circ$
counterpart. $\Theta$ thus uses frames up to a fixed lag, not the whole
trajectory; the overlap-weighted NCC sum in
\S\ref{sec:method_ncc} ranks the surviving wall families (mechanism
visualized in the supplement).

\subsection{Masked NCC and Sparse Observations}
\label{sec:method_ncc}
For a candidate heading and window, PadLoc computes Padfield's masked
NCC~\cite{padfield2012masked} between the premasked reference $R$ (mask $M_R$)
and the window target $T_w^b$ (graded mask $M_w^b$, \S\ref{sec:method_raster};
premasked $\tilde T_w^b=M_w^b\odot T_w^b$).
Writing $a\star c$ for the
cross-correlation $\mathcal F^{-1}\{\mathcal F(a)\,\overline{\mathcal F(c)}\}$
(one FFT pair per term), the field at shift $p$ is
\begin{equation}
  \mathrm{NCC}_w^b(p)=
  \frac{(R\star \tilde T_w^b)(p)
        -\dfrac{(R\star M_w^b)(p)\,(M_R\star \tilde T_w^b)(p)}{N(p)}}
       {\sqrt{D_1(p)\,D_2(p)}},
  \label{eq:masked_ncc}
\end{equation}
with the weighted overlap mass and the two per-image variances
\begin{align}
  N(p)   &=(M_R\star M_w^b)(p), \nonumber\\
  D_1(p) &=(R^2\star M_w^b)(p)-\frac{(R\star M_w^b)(p)^2}{N(p)},
  \label{eq:masked_ncc_terms}\\
  D_2(p) &=\bigl(M_R\star (M_w^b\odot (T_w^b)^2)\bigr)(p)-\frac{(M_R\star \tilde T_w^b)(p)^2}{N(p)}. \nonumber
\end{align}
Every term is one FFT correlation, so the field costs
$\mathcal{O}(HW\log HW)$. The overlap mass $N(p)$ restricts the means and
variances to $M_R\cap(M_w^b{+}p)$: observed-free zeros contribute, unobserved
cells drop out, and graded wall weights dominate the statistic. Shifts with
overlap mass $N(p)/\max_{p'}N(p')<0.7$ are rejected; all surviving cells store the
masked NCC evidence field
\begin{equation}
  F_w^b(p)=\mathrm{NCC}_w^b(p),
  \label{eq:evidence_field}
\end{equation}
a ranking and curvature surface rather than a calibrated probability density.

Heading selection is the one place an overlap weight enters. PadLoc centres each
field by its valid-cell mean and reweights it by
\begin{equation}
  \alpha_w^b(p)=\left(N(p)/|M_w^b|\right)^{0.70},
  \label{eq:overlap_weight}
\end{equation}
ranking each heading by the peak of
$\sum_w \alpha_w^b\,(\mathrm{NCC}_w^b-\overline{\mathrm{NCC}}_w^b)$. Here $|M_w^b|=\sum_p M_w^b(p)$ is total mask mass. This weight
is used only for cross-heading choice; the back-end receives the unweighted
field. The exponent $0.70$ is selected by its own sweep and is unrelated to
the $0.7$ overlap-rejection threshold above; the near-equal values are
coincidental.

From each field we extract a sparse observation
$Z_w^b=\{(\lambda_i,\mu_i,\Sigma_i)\}_{i=1}^{K}$. The locations
$\mu_i$ are the top-$K$ integer-cell maxima after non-maximum suppression
(no sub-cell interpolation), the weights $\lambda_i$ are softmax-normalized
peak scores, and $\Sigma_i=-H_i^{-1}\delta^2$ comes from the $3{\times}3$
finite-difference Hessian $H_i$ of the field at the peak, with an isotropic
fallback ($5$-cell standard deviation) at the crop boundary or when $H_i$
is not negative-definite.

\subsection{The \textsc{cons} Back-End}
\label{sec:method_backend}

\begin{figure*}[t]
  \centering
  \includegraphics[width=\textwidth]{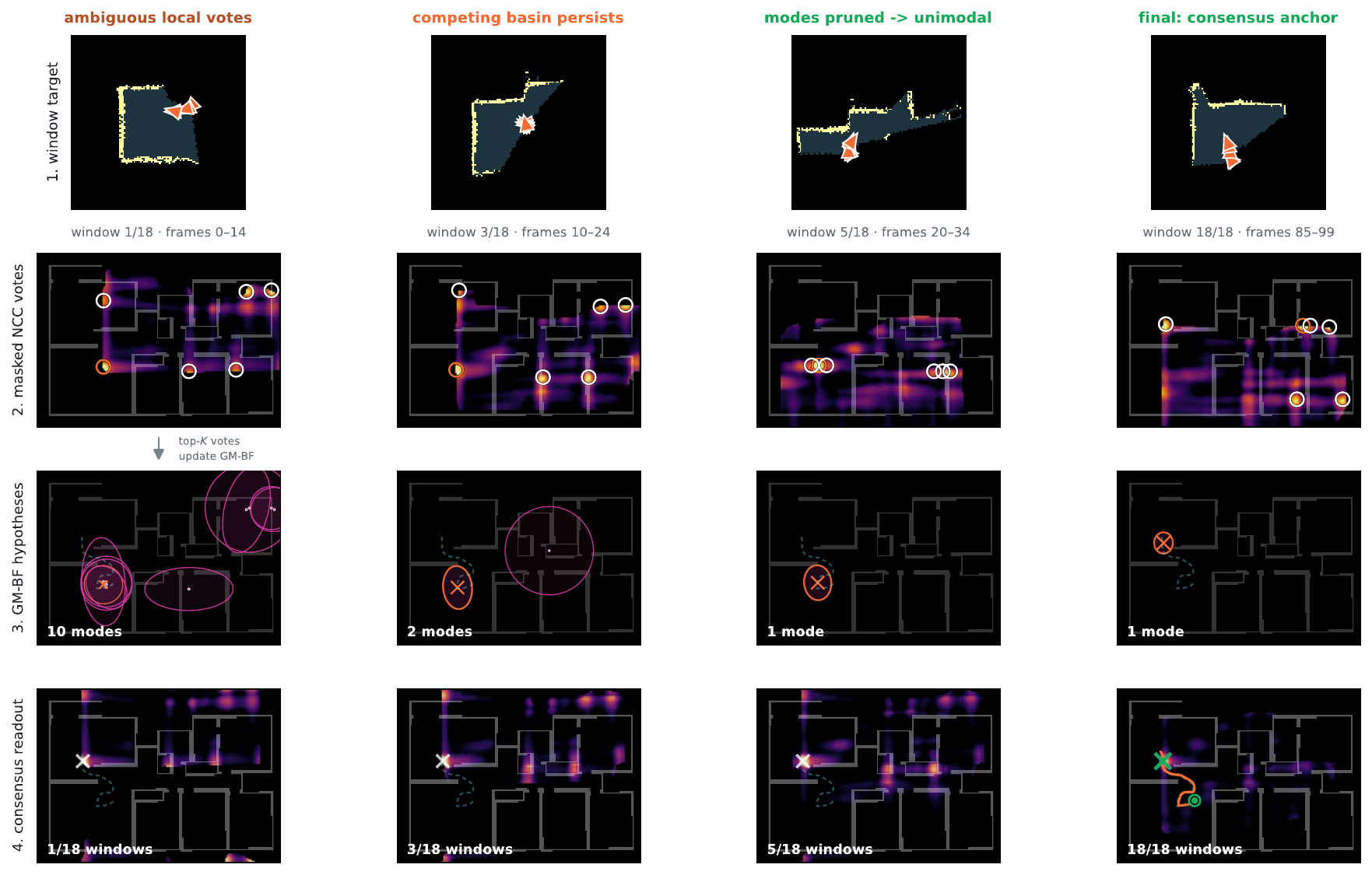}
  \caption{\textbf{Sliding-window matching and consensus decoding
  (\textsc{cons}).} Columns show one trajectory from early ambiguity to the
  final anchor. Rows show the accumulated target, masked-NCC top-$K$ votes,
  GM-BF hypotheses, and the cumulative consensus readout.}
  \label{fig:cons_walkthrough}
\end{figure*}

The \textsc{cons} decoder has two stages (Fig.~\ref{fig:cons_walkthrough}), split
by what each can trust: the GM-BF combines each window's top-$K$ peaks with
odometry to place the windows \emph{relative} to each other, correcting drift,
and the consensus sum adds the aligned raw fields to fix one \emph{absolute}
anchor. The filter alone drifts and can jump between basins; a raw sum alone
cannot align windows blurred by odometry error. First, a bounded
Gaussian mixture $\mathcal G_w=\{(W_h,m_h,P_h)\}_h$ tracks the absolute map
position of the current window centroid. Each window step shifts every mean
by the odometry centroid displacement $c_w-c_{w-1}$
and inflates every covariance by an isotropic process term
\begin{equation}
  Q=\bigl(\sigma\,\sqrt{s}\,\delta\bigr)^2\,\mathbf I,
  \label{eq:process_noise}
\end{equation}
where $\sigma{=}0.5$ is the per-frame motion-noise scale in cells, $\delta$
the grid resolution, and $s$ the inter-window stride in frames; added at
every window step, $Q$ grows linearly with elapsed frames (a random walk).
Predicted components are updated by Gaussian
products with the observation components in $Z_w^b$, pruned, moment-matched
when close, and capped at $K_{\max}=10$.
Before update, an observation is accepted only if it passes a
squared-Mahalanobis gate against the dominant (MAP, maximum a
posteriori) component $(m_\star^-,P_\star^-)$:
\begin{equation}
  (\mu_i-m_\star^-)^\top (P_\star^-)^{-1}(\mu_i-m_\star^-)\le16 .
  \label{eq:gmbf_gate}
\end{equation}
The wide gate uses $P_\star^-$ rather than $P_\star^-+\Sigma_i$, so broad
wrong-basin NCC peaks do not become easier to accept when their curvature is
flat. The mixture starts from the first window's top-$K$ peaks. Dominant-mode
gating can reject a correct alternative after a wrong lock. Covariance
inflation may later admit it, but recovery is not guaranteed. Component
birth is applied only to observations that pass the gate (supplement).

Second, the MAP sequence sets only the relative cyclic shifts before summing,
never the absolute position. The \emph{heal} replaces a MAP step by the odometry
step when the two differ by more than $\tau=\SI{0.3}{\meter}$:
\begin{equation}
  r_w=r_{w-1}+
  \begin{cases}
    \Delta c_w, & \|\Delta\hat z_w-\Delta c_w\|>\tau,\\
    \Delta\hat z_w, & \text{otherwise}.
  \end{cases}
  \label{eq:heal}
\end{equation}
Each raw field is then cyclically shifted into the last-window frame and summed:
\begin{align}
  S(p)&=\sum_{w=1}^{N}\mathrm{shift}(\tilde F_w,-o_w)(p), \\
  C(p)&=\sum_{w=1}^{N}\mathrm{shift}(V_w,-o_w)(p),
  \label{eq:consensus_sum}
\end{align}
where $V_w$ is the binary support mask of field $F_w$ (distinct from
the weighted target mask), and
$\tilde F_w=F_w-\overline{F_w[V_w]}$ centres each field over its support, preventing coverage bias from dominating agreement. Here $\mathrm{shift}(\cdot,-o_w)$
is a cyclic shift by the integer cell offset
$o_w=g_w-g_N$ between the grid cells $g_w$ of the $r_w$, and the count $C(p)$ is an
eligibility gate requiring $C(p)\ge\max(1,\mathrm{round}(N/2))$. As every window votes on the same frame-zero anchor, both
sums are kept running: a window is shifted by $-g_w$ and added once, and a
read shifts the sums by $g_N$.
The consensus peak $p^\star$ places the last-window centroid; subtracting
its cumulative odometry $c_N$ gives the frame-zero anchor
$a=p^\star-c_N$, from which all poses reconstruct as $\hat x_t=a+c_t$.
Restricting the heading read and this vote to
$[w{-}\mathrm{back},\,w{+}L]$ emits a pose for window $w$ once window
$w{+}L$ is acquired, i.e.\ at lag $L$; with unbounded past the last-window
pose equals the offline decode (\S\ref{sec:exp_setup}). Streamed poses are
reported at the window's mid-frame centroid.

The default configuration applies the tracker's causal per-window heading and
picks the wall family by the overlap-weighted fused sum of
Eq.~\eqref{eq:overlap_weight}; for long real-building trajectories an
optional repick runs the readout per $\theta_b$ and keeps the largest
consensus peak.

The fusion window length $W$ and stride $S$ couple the back-end to the
front-end's temporal structure. With per-frame depth the stride sets the
readout rate: $W{=}S{=}5$ emits one pose per five frames (\SI{2}{\hertz} on
Gibson), fuses every frame into exactly one window, and bounds the drift
inside each (\S\ref{sec:exp_ablation}). A front-end that instead reconstructs
in non-overlapping batches of $B$ frames (e.g.\ MapAnything's multiview
windows) requires $W{=}S{=}B$, so each fused target draws from a single
coherent reconstruction rather than straddling two (\S\ref{sec:exp_hilti}).

\section{Experiments}
\label{sec:experiments}

\subsection{Setup}
\label{sec:exp_setup}
We evaluate on the public \texttt{gibson\_t} split from
F3Loc~\cite{chen2024f3loc} over the Gibson environments~\cite{xia2018gibson}:
9 unseen scenes and 4045 frames at \SI{10}{\hertz}. Trajectories are re-chunked at
$T\in\{15,20,35,50,100\}$, matching UnLoc's denominators
($268/200/111/78/37$ chunks). The metric is last-10 SR@1\,m unless
otherwise noted: a chunk succeeds iff all of the last 10 frames are within
threshold. Because RMSE conditioned on each method's own
successes does not rank precision across methods, we also report the
\emph{all-trial} median last-10 RMSE (failures included) and \emph{Success
weighted by Accuracy}, $\mathrm{SWA}=\frac{1}{N}\sum_i s_i\,(1-e_i/\tau)$
--- an SPL-style score~\cite{anderson2018spl} over trials $i$ with success
$s_i$ at $\tau{=}\SI{1}{\meter}$ and last-10 RMSE $e_i$ (no established
localization variant exists, so we state the definition; Table~\ref{tab:swa}). Gibson and Bonn chunk trials reset
state between chunks; Hilti uses full runs.
In the Gibson comparisons, every method consumes
the same GT-derived relative transitions (F3Loc's transition step, UnLoc's
motion model, PadLoc's odometry); noise experiments perturb them with
identically seeded noise in each pipeline.

We keep latency regimes separate. \emph{Streaming} methods emit poses
online: F3Loc and UnLoc-\textsc{fwd} one per frame, PadLoc one per $5$-frame
window, first after frame~5. Our headline metric scores each of the last ten
frames with the latest pose emitted by then, propagated by odometry and never revised
(\emph{PadLoc frozen}). The \emph{chunk-end} readout instead scores the last
ten frames from the final estimate carried back by odometry (\emph{PadLoc
delayed}; supplement, with the timing details). \emph{Batched} methods need a
completed trajectory or an offline post-process: \emph{accum} builds a
whole-chunk target, and UnLoc refined is UnLoc's released postprocess, which
shifts its chunk-end trajectory by a bounded SE(2) fit of depth to the plan.

Depth sources: GT ray casts, F3Loc \texttt{mono}/\texttt{comp\_s}, UnLoc
DA-V2 $(\mu,\hat b)$, real LiDAR/RGB-D on Bonn. F3Loc and UnLoc comparisons use identical
cached depth inputs; the quoted Z-FLoc comparison shares the MapAnything
front-end, not its exact reconstructions.

\subsection{Matched-Depth Back-End Comparisons}
\label{sec:exp_backend}
Table~\ref{tab:backend_matched_depth} fixes each depth source and swaps
only the localization back-end. PadLoc \emph{frozen} is compared to
forward filters; PadLoc \textsc{accum} is compared to batched
whole-trajectory matchers or refinements. The UnLoc-depth block is
the controlled head-to-head. Under the chunk-end readout UnLoc-\textsc{fwd}
equals UnLoc refined in SR at every $T$ (supplement): the refinement gain UnLoc reports
($0.660{\to}0.810$ at $T{=}15$ here) comes entirely from switching the readout
from per-frame to chunk-end; the optimization alters no success flag. PadLoc frozen
exceeds UnLoc-\textsc{fwd} at every $T$ and the refinement from $T{=}50$ on
(tie at $T{=}35$); at $T{\le}20$ PadLoc accum trails the refinement.

\begin{table}[t]
\centering
\caption{\textbf{Matched-depth back-end comparison} on Gibson: last-10 SR@1\,m
($\uparrow$) and, following UnLoc's protocol, the mean last-10-frame RMSE over
each method's own successful trials (m, $\downarrow$; not a precision
ranking, Tab.~\ref{tab:swa}), pooled over 9 scenes. Each
F3Loc/UnLoc block fixes the depth input; F3Loc filter and UnLoc fwd use the
per-frame, PadLoc the frozen readout (\S\ref{sec:exp_setup}; chunk-end: supplement). Bold: best SR per
comparison (frozen vs filter, accum vs batched or refined rows), ties
bold. Z-FLoc quoted from its paper (`--').}
\label{tab:backend_matched_depth}
\scriptsize
\setlength{\tabcolsep}{1.5pt}
\begin{tabular}{l cc cc cc cc cc}
\toprule
 & \multicolumn{2}{c}{$T{=}15$} & \multicolumn{2}{c}{$T{=}20$} & \multicolumn{2}{c}{$T{=}35$} & \multicolumn{2}{c}{$T{=}50$} & \multicolumn{2}{c}{$T{=}100$} \\
\cmidrule(lr){2-3}\cmidrule(lr){4-5}\cmidrule(lr){6-7}\cmidrule(lr){8-9}\cmidrule(lr){10-11}
 & SR$\uparrow$ & R$\downarrow$ & SR$\uparrow$ & R$\downarrow$ & SR$\uparrow$ & R$\downarrow$ & SR$\uparrow$ & R$\downarrow$ & SR$\uparrow$ & R$\downarrow$ \\
\midrule
\multicolumn{11}{l}{\textit{F3Loc \texttt{mono} depth}} \\
F3Loc filter  & 0.291 & 0.27 & 0.345 & 0.23 & 0.550 & 0.21 & 0.705 & 0.18 & 0.892 & 0.18 \\
PadLoc frozen & \textbf{0.496} & 0.29 & \textbf{0.585} & 0.28 & \textbf{0.766} & 0.27 & \textbf{0.821} & 0.24 & \textbf{0.946} & 0.28 \\
\midrule
\multicolumn{11}{l}{\textit{F3Loc \texttt{comp\_s} depth}} \\
F3Loc filter  & 0.366 & 0.21 & 0.410 & 0.18 & 0.622 & 0.16 & 0.795 & 0.14 & \textbf{0.946} & 0.15 \\
PadLoc frozen & \textbf{0.563} & 0.25 & \textbf{0.685} & 0.25 & \textbf{0.883} & 0.22 & \textbf{0.949} & 0.22 & \textbf{0.946} & 0.22 \\
\midrule
\multicolumn{11}{l}{\textit{UnLoc $\mu$ depth}} \\
UnLoc fwd     & 0.660 & 0.13 & 0.710 & 0.14 & 0.874 & 0.12 & 0.923 & 0.14 & 0.973 & 0.16 \\
UnLoc refined & \textbf{0.810} & 0.17 & \textbf{0.865} & 0.17 & 0.919 & 0.17 & 0.949 & 0.17 & 0.973 & 0.20 \\
PadLoc frozen & \textbf{0.731} & 0.18 & \textbf{0.840} & 0.17 & \textbf{0.919} & 0.16 & \textbf{1.000} & 0.16 & \textbf{1.000} & 0.19 \\
PadLoc accum  & 0.795 & 0.16 & 0.860 & 0.17 & \textbf{0.946} & 0.14 & \textbf{1.000} & 0.14 & \textbf{1.000} & 0.15 \\
\midrule
\multicolumn{11}{l}{\textit{MapAnything depth, batched}} \\
Z-FLoc        & 0.667 & -- & 0.714 & -- & 0.865 & -- & 0.910 & -- & 0.995 & -- \\
PadLoc accum  & \textbf{0.787} & 0.24 & \textbf{0.825} & 0.22 & \textbf{0.928} & 0.18 & \textbf{1.000} & 0.16 & \textbf{1.000} & 0.14 \\
\bottomrule
\end{tabular}
\end{table}

PadLoc frozen exceeds the F3Loc filter by $+5$--$24$\,pp
(mono) and $+15$--$28$\,pp (\texttt{comp\_s}; tie at $T{=}100$); under the chunk-end readout it
leads or ties in every cell (supplement). PadLoc accum exceeds Z-FLoc's published
results at every $T$. Short-horizon failures are family flips or
early-window misses: a wrong family ($\ge45^\circ$) accounts for $37/72$ of the $T{=}15$ failures
on UnLoc depth and never coincides with a success; with the right family
the heading error is $1.3$--$2.1^\circ$ median (supplement).

Table~\ref{tab:swa} adds the unconditional view on the UnLoc-depth block.
At $T{\ge}35$ UnLoc-fwd is the more precise estimator (all-trial median
$0.11$--$0.12$ vs $0.15$--$0.19$\,m) and leads SWA at $T{=}100$;
PadLoc frozen leads UnLoc-fwd at every other $T$, accum beats offline refinement from
$T{=}35$ on, and
at \SI{5}{\centi\meter} the frozen configuration exceeds UnLoc-fwd at
every $T$. This resolution change also changes
the process noise in metres (Eq.~\eqref{eq:process_noise}); it is a
configuration comparison, not an isolated test of quantization.

\begin{table}[t]
\centering
\caption{\textbf{Unconditional metrics} on the Gibson UnLoc-$\mu$ block of
Tab.~\ref{tab:backend_matched_depth}: SWA $\uparrow$\,/\,all-trial median
last-10 RMSE (m) $\downarrow$; bold = best SWA per column (Bonn: supplement).}
\label{tab:swa}
\footnotesize
\setlength{\tabcolsep}{2.2pt}
\begin{tabular}{l ccccc}
\toprule
 & $T{=}15$ & $T{=}20$ & $T{=}35$ & $T{=}50$ & $T{=}100$ \\
\midrule
UnLoc fwd     & .573/.17 & .610/.14 & .771/.11 & .798/.11 & .822/.12 \\
UnLoc refined & \textbf{.670}/.18 & \textbf{.717}/.17 & .765/.16 & .791/.14 & .783/.17 \\
PadLoc frozen & .601/.20 & .695/.17 & .774/.15 & .839/.15 & .811/.19 \\
PadLoc accum  & .667/.18 & .715/.18 & \textbf{.809}/.14 & \textbf{.856}/.13 & .846/.14 \\
PadLoc frozen, \SI{5}{\centi\meter} & .624/.18 & .694/.15 & \textbf{.809}/.12 & .849/.11 & \textbf{.887}/.11 \\
\bottomrule
\end{tabular}
\end{table}

\subsection{Component Ablation}
\label{sec:exp_ablation}
Table~\ref{tab:ablation} isolates each component under the same frozen
readout (EMA heading tracker), on UnLoc $\mu$ depth pooled over the 9
Gibson scenes, and repeats the build-up under real drift (bottom panel).
\emph{Accumulation is the dominant back-end lever}: a per-frame NCC argmax
localizes almost nothing ($0.08$--$0.22$ SR@1\,m), and one
aggregate-then-match step matches or exceeds the full default at every $T$.

\begin{table}[t]
\centering
\caption{\textbf{Component ablation} (last-10 SR@1\,m $\uparrow$ / median last-10
RMSE (m) $\downarrow$, per-chunk reset; consensus rows frozen). Top:
Gibson build-up and leave-one-out (UnLoc $\mu$ depth, $n{=}268/111/37$; full
config bold). Bottom: the build-up under real drift (Bonn \texttt{static\_0},
D455$+$VO, $n{=}59/29/19$). Accumulation-alone is the offline
aggregate-then-match reference: the frozen decoder is within one chunk of it
at $T{\ge}35$, while under drift the gate keeps the readout near the
reference (row~3 vs~4) and above the smeared batch target.}
\label{tab:ablation}
\footnotesize
\setlength{\tabcolsep}{3pt}
\begin{tabular}{l cc cc cc}
\toprule
 & \multicolumn{2}{c}{$T{=}15$} & \multicolumn{2}{c}{$T{=}35$} & \multicolumn{2}{c}{$T{=}100$} \\
\cmidrule(lr){2-3}\cmidrule(lr){4-5}\cmidrule(lr){6-7}
 & SR & R & SR & R & SR & R \\
\midrule
\multicolumn{7}{l}{\textit{back-end build-up}} \\
per-frame NCC argmax            & 0.108 & 3.11 & 0.081 & 3.21 & 0.216 & 2.37 \\
\;$+$ accumulation (argmax)     & 0.817 & 0.17 & 0.928 & 0.14 & 1.000 & 0.13 \\
\;$+$ consensus, gate/heal off  & 0.687 & 0.21 & 0.856 & 0.15 & 0.919 & 0.14 \\
\;$+$ GM-BF gate \& heal        & \textbf{0.731} & 0.20 & \textbf{0.919} & 0.15 & \textbf{1.000} & 0.19 \\
\midrule
\multicolumn{7}{l}{\textit{leave-one-out from full}} \\
$-$ $\theta$-overlap weight     & 0.716 & 0.20 & 0.910 & 0.15 & 1.000 & 0.19 \\
$-$ $\hat b$ column drop        & 0.705 & 0.23 & 0.910 & 0.14 & 1.000 & 0.15 \\
$-$ consensus heal              & 0.724 & 0.21 & 0.892 & 0.15 & 0.973 & 0.19 \\
\midrule
\multicolumn{7}{l}{\textit{build-up under drift (Bonn D455$+$VO)}} \\
 & \multicolumn{2}{c}{$T{=}150$} & \multicolumn{2}{c}{$T{=}300$} & \multicolumn{2}{c}{$T{=}450$} \\
\cmidrule(lr){2-3}\cmidrule(lr){4-5}\cmidrule(lr){6-7}
per-frame NCC argmax            & 0.153 & 7.7\phantom{0} & 0.069 & 7.7\phantom{0} & 0.105 & 10.3\phantom{0} \\
\;$+$ accumulation (batched)    & 0.729 & 0.21 & 0.862 & 0.20 & 0.789 & 0.34 \\
\;$+$ consensus, gate off       & 0.525 & 0.42 & 0.552 & 0.29 & 0.368 & 3.4\phantom{0} \\
\;$+$ GM-BF gate \& heal        & \textbf{0.780} & 0.20 & \textbf{0.966} & 0.22 & \textbf{0.842} & 0.31 \\
\bottomrule
\end{tabular}
\end{table}

These frozen results use every past window (no forgetting),
$K{=}10$ hypotheses, and the running consensus sum.
The gate and heal matter mainly where motion drifts: removing both costs
$6.3/8.1$\,pp at $T{=}35/100$; the heal alone costs $2.7/2.7$\,pp. The remaining refinements help at
short and mid $T$ and vanish by $T{=}100$: the $\hat b$ column drop ($+2.6/{+}0.9$\,pp at $T{=}15/35$)
and the $\theta$-overlap weight ($+1.5$\,pp at $T{=}15$)---hence reading the
ablation across $T$.

Window/stride and raster-constant sweeps (supplement) show plateaus at
$T{\ge}35$, where the deployed $5/5$ is within one chunk of the best cell (tied at $T{=}100$).

The bottom panel of Table~\ref{tab:ablation} separates the components under real
drift. The association gate becomes decisive---held open, the readout
collapses from $0.842$ to $0.368$ SR@1\,m at $T{=}450$, the largest single
effect in either dataset, as ORB-SLAM3 relocalization jumps flood the filter
with wrong-basin observations. The frozen decode exceeds the
batch reference at every $T$ ($0.842$ vs $0.789$ at $T{=}450$): the whole-trajectory batched
target \emph{smears} under accumulated VO drift, while windowing bounds how
much drift enters any single correlation (the deployed $5/5$ beats $15/5$ by
up to $14$\,pp); only Hilti's batched reconstruction forces
$W{=}S{=}B$ (\S\ref{sec:exp_hilti}).

\paragraph{Bounded lag.}
On the Bonn \texttt{static\_0} \emph{drifty-VO} arm (clean
LiDAR\,\SI{106}{\degree} depth, drifty VO motion; supplement), the
$L{=}0$ window readout has a two-frame acquisition delay and steady-state
$p90$ error of \SI{0.76}{\meter} at $T{=}450$ (\SI{0.53}{\meter} for the offline $L{=}\infty$
reference). The textbook gate $P_\star^-{+}\Sigma_i$ performs alike here
($0.931/0.947$ for both at $T{=}300/450$; supplement).

\subsection{Real-Building Modularity on Bonn}
\label{sec:exp_bonn}
We evaluate the unchanged matcher and online \textsc{cons} back-end on Bonn
LiLocBench~\cite{trekel2025liloc} (AprilTag~\cite{olson2011apriltag} ground
truth), varying only off-the-shelf front-ends. Motion is wheel odometry or
monocular ORB-SLAM3~\cite{campos2021orbslam3} VO; depth is 2D LiDAR, D455
RGB-D, or---for the UnLoc head-to-head---a \emph{monocular structural depth}
pairing UnLoc's learned per-column depth with a Z-FLoc-style
farthest-return readout, so both methods localize from the same monocular
input. ORB-SLAM3 VO includes relocalization jumps, but the GM-BF gate and
the heal absorb most of them: VO trails wheel odometry by a
single chunk at $T{=}450$.
Across front-ends on \texttt{static\_0} ($T{=}300/450$; full table in the
supplement): full-FOV LiDAR reaches $1.000/1.000$ SR@1\,m under wheel
odometry (median RMSE \SIrange{9}{10}{\centi\meter}) and $1.000/0.947$ under
VO; narrow-FOV \SI{106}{\degree} LiDAR $0.931/0.947$; D455 RGB-D
$0.966/0.842$. We set $\sigma{=}1.0$ for clean LiDAR and $\sigma{=}0.5$ for
noisier depth; on RGB-D this moves one chunk ($0.842$ vs $0.789$ at
$T{=}450$). At $T{=}450$ RGB-D trails the LiDAR rows (${\ge}0.947$),
consistent with sensitivity to depth and motion quality.

\paragraph{Matched against UnLoc on real data.}
Holding monocular structural depth and the camera-built structural map (below) fixed, we
swap only the back-end across all four public Bonn sequences
(Table~\ref{tab:bonn_unloc}); UnLoc uses its learned per-column uncertainty
$\hat b$. Under deployable \emph{wheel} odometry, PadLoc beats UnLoc at every
pooled $T$ ($0.56/0.71/0.79$ vs $0.30/0.49/0.46$) and in 11 of 12
per-sequence cells, with one tie. The gain reflects representation as well
as time horizon: PadLoc registers the depth as 2D wall geometry and fuses each
window before a single match, whereas UnLoc's DeSDF likelihood scores $16$
rays per frame. Fed only UnLoc's own $16$ rays, PadLoc
reaches $0.47/0.67/0.77$ (Table~\ref{tab:bonn_unloc}; GT motion
$0.62/0.82/0.91$): $2$--$9$\,pp below the full profile and $+17$ to
$+30$\,pp above UnLoc. With ground-truth motion PadLoc is
stronger still (pooled $0.65/0.83/0.89$, $+35$ to $+43$\,pp). The unconditional metrics
widen the gap (supplement): PadLoc's pooled all-trial median
error stays sub-metre under wheel odometry
($0.46/0.25/\SI{0.24}{\meter}$) while UnLoc's sub-$50\%$ SR leaves it at
$5.3/1.0/\SI{3.5}{\meter}$; wheel SWA leads at every $T$
($0.43/0.55/0.62$ vs $0.23/0.36/0.35$), GT motion more still
($0.49/0.63/0.67$).

\begin{table}[t]
\centering
\caption{\textbf{PadLoc vs UnLoc on Bonn} (monocular structural depth and
camera-built structural map held fixed; UnLoc with its own $\hat b$). Last-10
SR@1\,m at $T{=}150/300/450$ per sequence and pooled; PadLoc frozen, UnLoc
per-frame readout (chunk-end: supplement);
PadLoc under \emph{wheel} odometry, UnLoc under its ground-truth
transitions. PadLoc wins 11 of 12 cells (tie: \texttt{dynamics\_0},
$T{=}450$, $n{=}5$). Pooled $n{=}172/84/56$. $16$ rays: PadLoc fed only the
$16$ DeSDF rays UnLoc's likelihood reads from the same profile.}
\label{tab:bonn_unloc}
\footnotesize
\setlength{\tabcolsep}{4pt}
\begin{tabular}{l rrr rrr}
\toprule
 & \multicolumn{3}{c}{UnLoc} & \multicolumn{3}{c}{PadLoc \textsc{cons}, wheel} \\
\cmidrule(lr){2-4}\cmidrule(lr){5-7}
scene & 150 & 300 & 450 & 150 & 300 & 450 \\
\midrule
\texttt{static\_0}      & 0.32 & 0.59 & 0.68 & \textbf{0.59} & \textbf{0.79} & \textbf{0.89} \\
\texttt{dynamics\_0}    & 0.33 & 0.43 & \textbf{0.60} & \textbf{0.67} & \textbf{0.71} & \textbf{0.60} \\
\texttt{lt\_changes\_0} & 0.21 & 0.43 & 0.21 & \textbf{0.60} & \textbf{0.71} & \textbf{0.79} \\
\texttt{lt\_ch\_dyn\_0} & 0.33 & 0.44 & 0.39 & \textbf{0.45} & \textbf{0.63} & \textbf{0.72} \\
\midrule
pooled SR@1\,m          & 0.30 & 0.49 & 0.46 & \textbf{0.56} & \textbf{0.71} & \textbf{0.79} \\
pooled, $16$ rays       &      &      &      & 0.47 & 0.67 & 0.77 \\
succ.\ med.\ RMSE (m)  & 0.19 & 0.25 & 0.25 & 0.19 & 0.20 & 0.20 \\
\bottomrule
\end{tabular}
\end{table}

\paragraph{Reference map: shipped vs structural.}
The benchmark (shipped) map is a 2D LiDAR scan at one height that records static
furniture (e.g.\ a kitchenette) as wall---exactly what our structural depth
omits. We therefore rebuild a furniture-free structural reference
from the three D455 cameras of the benchmark's \emph{mapping} run (Mask2Former
wall/door/window masking; no test-sequence information) and report both methods on
both maps. On the \emph{shipped} map, under the identical configuration as
Table~\ref{tab:bonn_unloc}, PadLoc leads UnLoc at every pooled $T$:
$0.49/0.62/0.73$ vs $0.37/0.52/0.57$ under wheel odometry and
$0.63/0.79/0.86$ with ground-truth motion. Which map is better depends on
the back-end: the walls-only rebuild helps aggregate-then-match (up to
$+10$\,pp) but \emph{costs} UnLoc's per-frame filter up to $11$\,pp---UnLoc's
best map is the shipped one---so we report both maps for both methods. The gap
is present on both maps and widens
on the structural map to the headline $+23$--$32$\,pp. Masking camera pixels
while keeping the LiDAR changes nothing ($0.400{\,=\,}0.400$ on
\texttt{dynamics\_0}, $T{=}150$); dropping the LiDAR-baked furniture is the
entire lever ($0.400{\to}0.600$; full matrix in the supplement).

\subsection{Large-Building Modularity on Hilti}
\label{sec:exp_hilti}
Bonn fixes the building at apartment scale and the front-end at per-frame depth. The
2026 Hilti-Trimble Localization
Challenge~\cite{helmberger2022hilti,centanni2026hiltitrimble} stresses both:
multi-hundred-metre
construction floors (up to $T{=}6124$ frames) localized from a camera-only fisheye
rig. The five ground-truth sessions span three physical floors---\texttt{floor\_1},
three sessions of \texttt{floor\_2}, and the \texttt{UG1} underground hall
(Table~\ref{tab:hilti}). Here too we assemble the front-end after Z-FLoc~\cite{zfloc2026}: depth from
MapAnything~\cite{keetha2026mapanything} multiview reconstruction (in
\emph{non-overlapping $30$-frame batches}) and an
OKVIS2-X~\cite{boche2025okvis2x} visual-inertial odometry prior we run on the
rig. Because MapAnything cannot ingest the raw fisheye, we rectify it to an
\SI{80}{\degree} pinhole---discarding most of the rig's wide-FOV advantage.
As the building is an order of magnitude larger than Bonn, we report
\emph{online} SR@\SI{2}{\meter}: the fraction of \emph{all} emitted window
poses within \SI{2}{\meter} over the full run, cold start included (the
challenge threshold; per-pose, not last-10).

We set $W{=}S{=}30$, MapAnything's batch, for every session
(\S\ref{sec:method_backend}): each fused target is then one complete
reconstruction, whereas windows that straddle batches or fuse only five frames
of one collapse on most sessions (supplement). The az55 and $+$seg front-ends are two alternatives,
each applied unchanged to all sessions, with ground truth used for evaluation only;
we do not evaluate a selector between them.
On three of the five sessions PadLoc then localizes at $0.90$--$0.98$. The other
two expose \emph{front-end} limits of different kinds: run\_1's
reconstruction is wall-like but as-built rather than as-drawn; UG1's depth
collapses to blobs with nothing to register (supplement)---the Bonn RGB-D
depth-quality bottleneck at building scale.

\paragraph{Multi-crop and wall-isolation front-end arms.}
Rectifying two additional $\pm55^\circ$ crops per fisheye (six views;
back-end unchanged) quantifies the lost field of view (Table~\ref{tab:hilti}, az55):
the weakest localizable session rises $0.456{\to}0.689$ and run\_2 improves,
while 05-05 and the strongest drop ($0.977{\to}0.710$) as coverage trades against
per-view quality. Adding learned wall isolation (SegFormer wall/door/window;
$+$seg) recovers UG1 outright, $0.156{\to}0.668$, improves the two
sessions that rely on free space ($0.920$ on 05-05, $0.689$ on run\_1), and lowers floor\_1, whose walls match the plan, from
$0.977$ to $0.718$. The uniform $+$seg variant therefore reaches $0.668$--$0.920$
across all five sessions with the back-end unchanged.

\paragraph{Classical control on the same front-end.}\looseness=-1
Replacing the back-end by a memoryless per-block $(x,y,\theta)$ argmax---the
same masked-NCC kernel and rasters, each $30$-frame block localized over a
dense $5^\circ$ sweep---collapses to $0.05$--$0.11$ SR@\SI{2}{\meter}
(median $15$--$18$\,m; wrong heading family on $66$--$77\%$ of blocks).
Using our pre-pass headings changes little ($0.05$--$0.15$): single blocks
are globally ambiguous, and PadLoc's $0.90$--$0.98$ comes from gated GM-BF
and consensus across blocks.

\begin{table}[t]
\centering
\caption{\textbf{Hilti large-building localization} (camera-only fisheye,
MapAnything depth, OKVIS2-X prior). Online SR@2\,m ($\uparrow$) for the
batch-matched $30/30$ window and front-end arms (az55: $\pm55^\circ$
side crops; $+$seg: learned wall isolation), back-end unchanged; each column
is one configuration applied to all sessions, none is selected per session;
mean error/score at $30/30$. \texttt{run\_2}, \texttt{05-05}, \texttt{run\_1}:
the three \texttt{floor\_2} sessions (2025-10-28 run~2, 2025-05-05,
2025-10-28 run~1).}
\label{tab:hilti}
\small
\setlength{\tabcolsep}{4.3pt}
\begin{tabular}{l r ccc c c}
\toprule
session & $T$ & \multicolumn{3}{c}{SR@2\,m} & mean & score \\
\cmidrule(lr){3-5}
 & & $30/30$ & az55 & $+$seg & (m) & \\
\midrule
floor\_1 & 3923 & 0.977 & 0.710 & 0.718 & 0.90 & 68.1 \\
run\_2   & 2569 & 0.907 & 0.953 & 0.895 & 0.68 & 74.3 \\
05-05    & 5594 & 0.898 & 0.813 & 0.920 & 1.11 & 62.3 \\
\midrule
run\_1   & 3067 & 0.456 & 0.689 & 0.689 & 2.68 & 32.3 \\
UG1      & 6124 & 0.156 & 0.000 & 0.668 & 5.62 & \phantom{0}9.9 \\
\bottomrule
\end{tabular}
\end{table}

\subsection{Complexity, Runtime, and Footprint}
\label{sec:exp_ondevice}

\begin{figure}[t]
  \centering
  \includegraphics[width=\columnwidth]{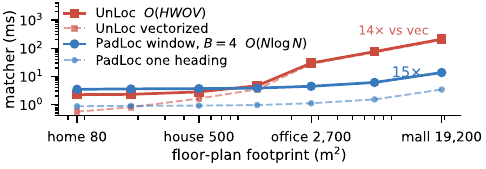}
  \caption{\textbf{Matcher cost vs.\ floor-plan size}
  (\SI{10}{\centi\meter} grid; RTX 3090). One UnLoc call scores all
  36 orientations per frame; a PadLoc window scores all $B{\approx}4$
  headings (solid; one heading dashed); complete back-end cost is given in
  the text. UnLoc grows $2.3\!\to\!\SI{209}{\milli\second}$; a PadLoc
  window grows $3.5\!\to\!\SI{13.7}{\milli\second}$ ($15\times$ at mall
  scale, $76\times$ per frame at stride $5$; UnLoc is cheaper at home scale).}
  \label{fig:runtime}
\end{figure}

With depth fixed, UnLoc scans the full DeSDF every frame over $O{=}36$
orientations and $V{=}11$ rays ($\bigO(HWOV)$), while PadLoc correlates one
fused target over $B{\approx}4$ headings per $S{=}5$-frame window with cached
reference FFTs ($\bigO(N\log N)$)---the Fig.~\ref{fig:runtime} gap. The
kernel takes \SIrange{5}{40}{\milli\second} per heading on iPhone
15/17~Pro at room-to-floor scale, rising to
\SIrange{176}{224}{\milli\second} at mall scale (supplement).

On Gibson / Bonn, PadLoc's back-end holds the cached FFTs of the
reference (\SIrange{1.2}{1.9}{\mega\byte} / \SI{9.9}{\mega\byte}; built
from the plan in ${\le}\SI{0.4}{\second}$), \SI{13}{\byte} per cell and
heading, and a \SI{560}{\byte} mixture, i.e.\ $\bigO(BHW)$ evidence state
for any number of windows under an unbounded past. A window costs one
shift-and-add per heading and a read one shift. UnLoc holds a float32
$(x,y,\theta)$ prior and posterior at its \SI{10}{\centi\meter} grid
(\SIrange{8}{44}{\mega\byte} / \SI{147}{\mega\byte} on the GPU) and a DeSDF
precomputed per map, grid and orientation count (\SI{4}{\minute} to
${\approx}\SI{77}{\minute}$ / ${\approx}\SI{44}{\minute}$ with its released
code; supplement).

End-to-end, the full back-end is ${\sim}\SI{40}{\milli\second}$ per emitted
window at Gibson scale on a laptop CPU (Apple M1~Max, no GPU):
${\sim}\SI{15}{\milli\second}$ for the $B{\approx}4$ masked-NCC matches and
${\sim}\SI{25}{\milli\second}$ for rasterization, pre-pass, and
GM-BF$+$consensus. At stride $S{=}5$ this is ${\sim}\SI{8}{\milli\second}$ per
frame (throughput, not reporting delay); the shared DA-V2 ViT-L depth
front-end costs \SI{114}{\milli\second}, so localization is not the bottleneck.

\section{Discussion and Conclusion}
\label{sec:discussion}

\paragraph{Limits.}\looseness=-1
Very short sparse-depth trajectories still suffer directed heading
ambiguity ($14$--$29\%$ wrong-family chunks at $T{=}15$: the free-space wedge is
too small to reject the $180^\circ$ mirror). Under high drift the filter can stay locked on a wrong basin; the consensus
peak-to-sidelobe ratio (PSR) flags this but does not relocalize.
The method assumes stable wall structure; furniture, renovations, or
long-term change make map quality decisive. The five Hilti sessions split
into three regimes: matching walls, as-built walls where free space
carries, and clutter where wall isolation recovers structure
(\S\ref{sec:exp_hilti}). No variant is best on every session, and we do not
evaluate a selector between them. Fixed parameters were set on Gibson; Bonn modularity
and lag sweeps concentrate on \texttt{static\_0} at small $n$.

\paragraph{Conclusion.}
PadLoc---masked correlative matching over dominant wall orientations,
consensus-decoded by a moving-pose GM-BF---matches or beats native back-ends
under matched depth and transfers across real buildings, with front-end knobs
set by rule rather than tuning. Code: \url{github.com/mbookhahn/padloc}.

\clearpage
{
    \small
    \bibliographystyle{ieeenat_fullname}
    \bibliography{bib/refs}
}

\clearpage
\appendix
\raggedbottom
\renewcommand{\thefigure}{S\arabic{figure}}
\renewcommand{\thetable}{S\arabic{table}}
\setcounter{figure}{0}
\setcounter{table}{0}
\makeatletter
\setlength{\@fptop}{0pt}
\setlength{\@fpbot}{0pt plus 1fil}
\setlength{\@fpsep}{12pt}
\makeatother

\begin{center}\large\bfseries Supplementary Material\end{center}
\noindent This appendix collects ablations and secondary results referenced
from the main paper; references of the form ``main paper'' point to the
sections and tables above.

\section{Observation semantics and reference blur}
\label{app:obs_semantics}
\begin{figure}[t]
  \centering
  \includegraphics[width=\columnwidth]{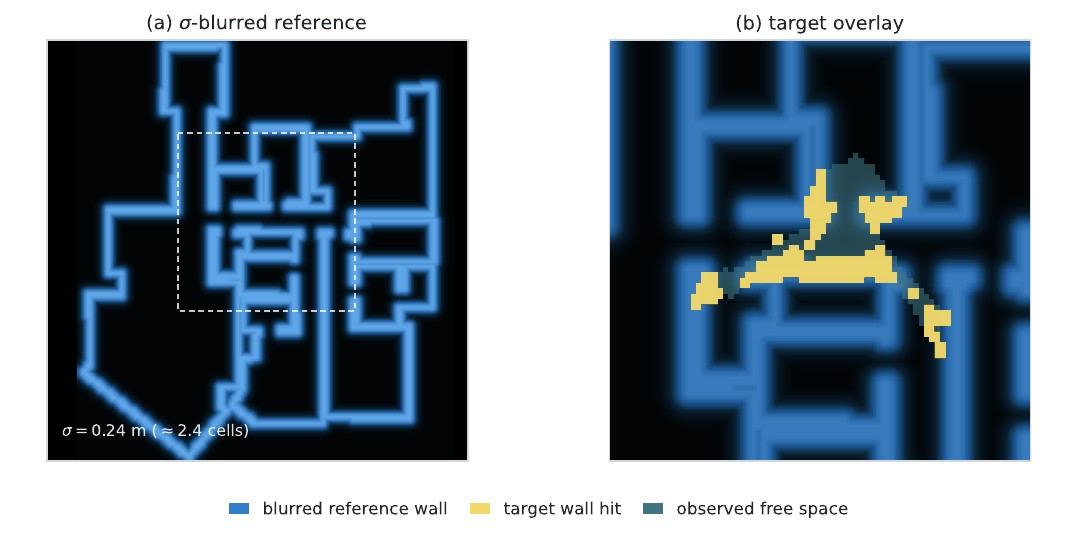}
  \caption{\textbf{Observation semantics on the blurred reference.}
  (a)~$\sigma$-blurred reference field. (b)~A window target overlaid:
  yellow wall hits register on the blue reference bands, teal cells are
  observed free space, and unobserved cells drop out of the masked NCC.}
  \label{fig:blur_supp}
\end{figure}

Figure~\ref{fig:blur_supp} details the observation raster sketched in the
main paper's teaser: wall hits carry mask weight $w_{\mathrm{wall}}$,
traversed cells become observed free space whose zeros \emph{penalize}
overlapping reference walls, and unobserved cells drop out of the masked
statistics entirely. Reference wall strokes are blurred to the scale of
depth noise ($\sigma{=}\SI{0.24}{\meter}$) so sensor wall hits register on
the reference bands.

\paragraph{Weighted masked NCC.}
Both arrays are premasked: $R=M_RR_{\mathrm{raw}}$ and
$\tilde T=MT$, with binary reference mask $M_R$, binary target $T$,
and $M\in\{0,1,w_{\mathrm{wall}}\}$; the kernel applies Padfield's
Algorithm~1 to these premasked arrays with the weighted target
second moment. Main-paper
Eqs.~(1)--(2) give the score, including the $M T^2$ term in $D_2$, the
exact weighted variance (applying Algorithm~1 unchanged to $\tilde T$ would
give $M^2T^2$).

Numerically, the code adds $10^{-3}$ to overlap mass, clips $D_1,D_2$
below at zero, and accepts normalization only when both exceed
$10^{-6}N$. It floors $\sqrt{D_1D_2}$ at $10^{-12}$ and assigns zero
otherwise; the overlap gate then excludes shifts below $0.7\max N$.
These safeguards accompany, rather than alter, the stated premasking
convention. All reported results use this score.

\begin{figure*}[t]
  \centering
  \includegraphics[width=\textwidth]{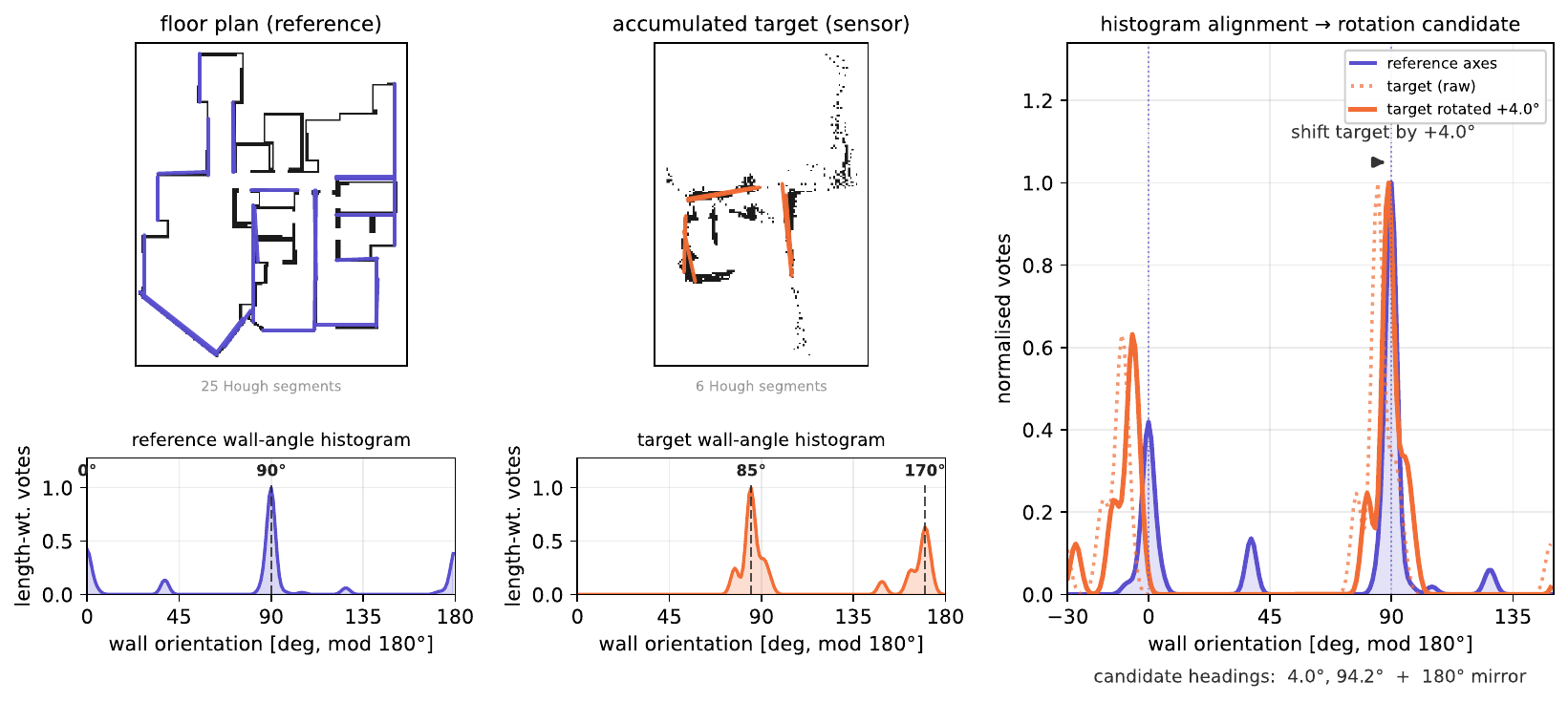}
  \caption{\textbf{Orientation pre-pass mechanism.} Hough wall segments from
  the reference and accumulated target scan are converted into length-weighted
  wall-angle histograms. Cross-correlation aligns the target histogram to the
  reference axes and yields the directed heading set plus $180^\circ$ mirrors
  (here $4.0^\circ/94.2^\circ$), which masked NCC scores downstream.}
  \label{fig:prepass_supp}
\end{figure*}

\section{Multi-hypothesis recall}
\label{app:topk}
\begin{figure}[t]
  \centering
  \begin{tikzpicture}
    \begin{axis}[width=0.98\columnwidth, height=0.60\columnwidth,
        xlabel={$K$ (NMS peaks kept)}, ylabel={cum.\ recall [\%]},
        xmode=log, log ticks with fixed point,
        xmin=0.93, xmax=21.5, ymin=55, ymax=103,
        xtick={1,3,5,10,20}, ytick={60,70,80,90,100},
        grid=both, grid style={black!12},
        tick label style={font=\scriptsize}, label style={font=\small},
        legend style={font=\scriptsize, at={(0.97,0.05)}, anchor=south east,
                      draw=none, fill=white, fill opacity=0.85, text opacity=1},
        legend cell align=left]
      \addplot+[thick, mark=*, mark size=1.5, color=black!60]
        coordinates {(1,97.3)(3,97.3)(5,97.3)(10,97.3)(20,97.3)};
      \addplot+[thick, mark=square*, mark size=1.5, color=teal!80!black]
        coordinates {(1,97.3)(3,100)(5,100)(10,100)(20,100)};
      \addplot+[thick, mark=triangle*, mark size=2.0, color=orange!85!black]
        coordinates {(1,86.5)(3,91.9)(5,97.3)(10,100)(20,100)};
      \addplot+[thick, mark=diamond*, mark size=2.0, color=red!75!black]
        coordinates {(1,62.2)(3,78.4)(5,78.4)(10,83.8)(20,86.5)};
      \legend{$\sigma{=}0$, $\sigma{=}0.5$, $\sigma{=}1.0$, $\sigma{=}2.0$}
      \draw[dashed, black!45] (axis cs:10,55) -- (axis cs:10,103)
        node[pos=0.97, left, font=\scriptsize, black!60] {deployed $K$};
    \end{axis}
  \end{tikzpicture}
  \caption{Cumulative top-$K$ NMS recall of the GT pose on the \emph{headline
  consensus decoder's} surface (moving-pose GM-BF $+$ consensus-sum,
  chunk-end readout, GT depth, $\SI{10}{\centi\meter}$ grid, $37$ \texttt{gibson\_t}
  chunks; GT-hit and NMS separation $\SI{0.30}{\meter}$,
  quantization-fair at $\SI{10}{\centi\meter}$). $\sigma$ is per-frame odometry
  noise: a seeded random walk of $\sigma\,$cm on each translation axis and
  $\sigma^{\circ}$ on heading per frame. At $\sigma{=}2.0$ the global argmax
  recovers GT in only $\SI{62.2}{\percent}$ of chunks, while the deployed
  $K{=}10$ (dashed) recovers $\SI{83.8}{\percent}$ (top-$20$: $\SI{86.5}{\percent}$); for
  $\sigma{\le}0.5$ the argmax misses at most one chunk (at $\sigma{=}0$ one chunk is missed at
  every $K$).}
  \label{fig:topk_rank}
\end{figure}
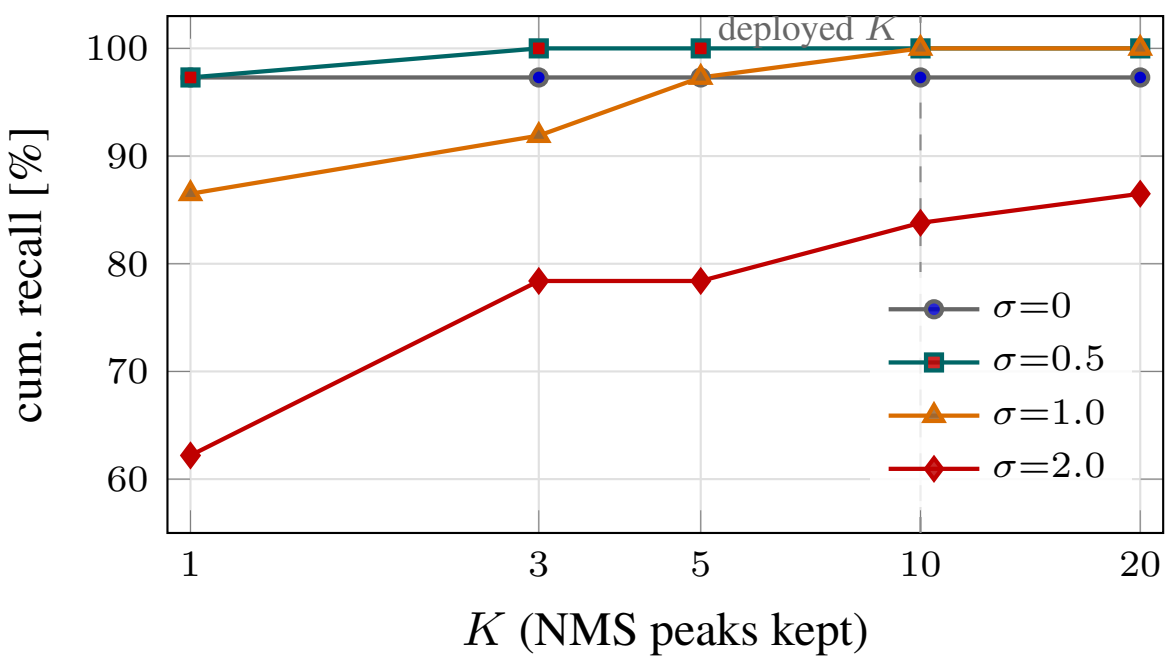

Figure~\ref{fig:topk_rank} examines residual multimodality under
increasing motion noise: the global argmax of the consensus surface
slips off ground truth on a growing fraction of chunks, while the true
pose remains among its top-$K$ NMS peaks more often. This is a diagnostic of the final consensus surface, not a
measurement of candidate recall in the individual input fields.

\section{Hyperparameters}
\label{app:hyperparams}
Table~\ref{tab:hyperparams} lists the principal matcher and back-end parameters, their
values, and how each was set. All fixed knobs were selected once on Gibson, so
we disclose this rather than claim a Gibson holdout. They are then never
re-tuned, never per scene; Bonn and Hilti provide cross-dataset holdout
evidence for those choices. Only the
front-end-determined group changes across datasets, each by a stated rule:
the fusion window follows the readout rate for per-frame depth and equals
the reconstruction batch for batched front-ends (main paper, Method), the motion-noise scale follows the
odometry/depth noise class, the reference blur follows the depth class
(learned depth \SI{0.24}{\meter}; physical range sensors
\SI{0.15}{\meter}), and the cell size matches the competitor grid (Gibson)
or the benchmark map's native resolution (Bonn).

\begin{table}[t]
\centering
\caption{\textbf{Principal hyperparameters.} Top: fixed across all
datasets. Bottom: front-end-determined, set by stated rules.}
\label{tab:hyperparams}
\scriptsize
\setlength{\tabcolsep}{2pt}
\begin{tabular}{@{}p{0.40\columnwidth}p{0.27\columnwidth}p{0.26\columnwidth}@{}}
\toprule
parameter & value & set by \\
\midrule
\multicolumn{3}{l}{\textit{fixed across all datasets}} \\
wall-hit mask weight $w_{\mathrm{wall}}$ & $55$ & Gibson \\
$\hat b$ column-drop threshold & \SI{0.30}{\meter} & Gibson \\
overlap-rejection threshold & $0.7$ & Gibson \\
heading-overlap exponent & $0.70$ & own sweep (unique opt.) \\
top-$K$ peaks / mixture cap & $10$ & Gibson \\
Mahalanobis gate (squared) & $16$ & Gibson \\
heal threshold $\tau$ & \SI{0.3}{\meter} & Gibson \\
consensus eligibility & $\operatorname{round}(N/2)$, min.\ 1 & Gibson \\
pre-pass scan / cadence & $5$\,fr / every $5$ & Gibson \\
heading-tracker EMA & $0.6$ & Gibson \\
\midrule
\multicolumn{3}{l}{\textit{BEV projection (walls by geometry, no classifier)}} \\
height band (floor-rel.) & \SIrange{0.3}{2.4}{\meter} & knee to doorway top \\
radius clip & \SI{8}{\meter} & front-end confidence range \\
azimuth bin / reduce & \SI{0.5}{\degree} / median & robust range per bearing \\
cluster gap / min pts & \SI{0.5}{\meter} / 3 & segment contiguity \\
accum support radius & \SI{0.2}{\meter} ($\ge$2 fr.) & cross-frame support \\
\midrule
\multicolumn{3}{l}{\textit{front-end-determined}} \\
fusion window $W/S$ & $5/5$; $30/30$ Hilti & readout rate; $=$ batch \\
motion noise $\sigma$ & $0.5$; $1.0$ LiDAR & odometry/depth noise class \\
reference blur $\sigma_{\mathrm{blur}}$ & \SI{0.24}{\meter}; \SI{0.15}{\meter} range sensors & depth-noise class \\
cell size $\delta$ & \SI{10}{\centi\meter}; \SI{5}{\centi\meter} Bonn & competitor grid / native map \\
\bottomrule
\end{tabular}
\end{table}

\paragraph{Sensitivity of $w_{\mathrm{wall}}$ and the $\hat b$ threshold.}
Table~\ref{tab:sens} sweeps the two remaining fixed constants on the
UnLoc-depth arm at every $T$. The wall-hit weight sits on a broad plateau: a $4\times$
range ($25$--$100$) moves pooled SR@1\,m by at most $3.7$\,pp at any $T$
(the published $55$ is the optimum at neither $T{=}15$ nor $T{=}35$). Across
$0.20$--$0.40$\,m the $\hat b$-drop threshold moves pooled SR by at most
$4.1$\,pp, about as much as removing the drop (off: $-2.6/-5.5/-0.9/-2.6/0$\,pp,
reproducing the main paper's leave-one-out row exactly, $0.705/0.910/1.000$;
$\tau{=}0.20$ equals off at $T{=}15$ and $50$). Running the pair the test-set
sweeps favoured with the re-summed consensus (rounded offsets,
Sec.~\ref{app:complexity}; $w_{\mathrm{wall}}{=}40$, $\tau{=}0.25$) changes
pooled SR by
$+1.1/-0.5/0/-1.3/0$\,pp ($+9/{-}6$ chunks at $T{=}15$): the published
constants are within a few chunks of a test-tuned configuration, not at
its optimum.

\begin{table}[t]
\centering
\caption{\textbf{Sensitivity sweeps} (UnLoc $\mu$ depth, 9 scenes, pooled
last-10 SR@1\,m, frozen readout). One constant is varied at a time from the
published setting (bold); last row: the pair the test-set sweeps favoured with the re-summed consensus (Sec.~\ref{app:complexity}).}
\label{tab:sens}
\footnotesize
\setlength{\tabcolsep}{2.5pt}
\begin{tabular}{l ccccc}
\toprule
setting & $T{=}15$ & $T{=}20$ & $T{=}35$ & $T{=}50$ & $T{=}100$ \\
\midrule
\textbf{published} ($w{=}55$, $\tau{=}0.30$) & \textbf{.731} & \textbf{.840} & \textbf{.919} & \textbf{1.000} & \textbf{1.000} \\
\midrule
$w_{\mathrm{wall}}{=}25$  & .746 & .835 & .910 & .974 & 1.000 \\
$w_{\mathrm{wall}}{=}40$  & .728 & .840 & .928 & 1.000 & 1.000 \\
$w_{\mathrm{wall}}{=}70$  & .713 & .835 & .937 & 1.000 & 1.000 \\
$w_{\mathrm{wall}}{=}100$ & .709 & .815 & .928 & .987 & 1.000 \\
\midrule
$\hat b$ cutoff $\tau{=}0.20$ & .705 & .815 & .937 & .974 & 1.000 \\
$\tau{=}0.25$ & .746 & .840 & .919 & 1.000 & 1.000 \\
$\tau{=}0.40$ & .713 & .830 & .937 & 1.000 & 1.000 \\
$\hat b$ drop off & .705 & .785 & .910 & .974 & 1.000 \\
\midrule
$w{=}40$, $\tau{=}0.25$ & .743 & .835 & .919 & .987 & 1.000 \\
\bottomrule
\end{tabular}
\end{table}

The fusion window is the main front-end-coupling choice. The full sweep
(Table~\ref{tab:ws_sweep}) shows a plateau at $T{\ge}35$ rather than a sharp optimum: SR
stays within $3.6$\,pp across all cells there, and under the frozen readout
the deployed $5/5$ is within one chunk of the best cell at $T{=}35$ ($15/5$,
$0.928$ vs $0.919$) and tied at $T{=}100$; at $T{=}15$ only the $5$-frame
windows serve every scored frame; longer windows back-fill the first scored
frames from their first output, and $15/5$ is a single window read at the
chunk end (parentheses). On the
high-drift Bonn VO arm $5/5$ is better than $15/5$ by up to $14$\,pp
(Sec.~\ref{app:bonn_ablation})---the drift-bound argument favours smaller
windows there---and we keep one global $5/5$ across all per-frame-depth
datasets, as it follows from the readout rate ($W$
set to the existing stride $S$, no further constant); the $15/5$ cells
remain in Table~\ref{tab:ws_sweep}.

\begin{table}[t]
\centering
\caption{\textbf{Fusion window/stride sweep on Gibson} (UnLoc $\mu$ depth,
9 scenes, last-10 SR@1\,m $\uparrow$ / median last-10 RMSE (m) $\downarrow$,
frozen readout; $n{=}268/111/37$; at $T{=}15$, windows longer than five frames emit their first pose after scoring starts and earlier scored frames take that pose ($15/5$: one window, frozen $=$ chunk-end), in parentheses). The operating point sits on a
plateau at $T{\ge}35$.}
\label{tab:ws_sweep}
\footnotesize
\setlength{\tabcolsep}{3pt}
\begin{tabular}{l cc cc cc}
\toprule
 & \multicolumn{2}{c}{$T{=}15$} & \multicolumn{2}{c}{$T{=}35$} & \multicolumn{2}{c}{$T{=}100$} \\
\cmidrule(lr){2-3}\cmidrule(lr){4-5}\cmidrule(lr){6-7}
$W/S$ & SR & R & SR & R & SR & R \\
\midrule
$5/1$   & 0.683 & 0.24 & 0.901 & 0.16 & 0.973 & 0.21 \\
$5/5$ (deployed) & 0.731 & 0.20 & 0.919 & 0.15 & 1.000 & 0.19 \\
$10/5$  & (0.769) & 0.20 & 0.910 & 0.15 & 1.000 & 0.17 \\
$10/10$ & (0.761) & 0.21 & 0.892 & 0.16 & 1.000 & 0.13 \\
$15/5$  & (0.817) & 0.17 & 0.928 & 0.16 & 1.000 & 0.19 \\
\bottomrule
\end{tabular}
\end{table}

\section{Confidence thresholding versus intensity weighting}
\label{app:bhat_weight}
The main paper (\S3) treats depth \emph{uncertainty} as \emph{membership}
rather than intensity --- for $\hat b$ the discriminative control is which
cells participate in the masked NCC, not how strongly each is weighted. The
mask does carry one fixed \emph{geometric} weight (wall hits over
observed-free cells, \S3), and the overlap mass $\alpha_w^b$ weights the
cross-heading selection; what the back-end field does \emph{not} use is a
per-cell \emph{depth-confidence} weight. Here we report the ablation that
motivates that membership choice.

An alternative is to weight each rasterized cell by the depth head's
confidence, down-weighting free-space cells with large aleatoric
uncertainty $\hat b$ within the weighted score. On
\texttt{gibson\_t} this is negative: against the same configuration without
any use of $\hat b$, pooled last-10 SR@1\,m (frozen readout) \emph{drops} by
$10$--$23$\,pp at $T{\le}50$ (both reach $1.000$ at $T{=}100$), and all $9$
scenes regress; weighting only the wall hits loses as much. This empirical result motivates the default column-drop rule, but
does not establish why weighting fails; score normalization and the
per-frame mean-1 weight normalization may interact with confidence
weighting. We therefore restrict the conclusion to the evaluated
configuration: thresholding membership performed better than the tested
intensity weighting.

\section{Gate-space relaxations (full)}
\label{app:gate}
The main paper adopts a $P_\mathrm{map}$-only association gate; the
textbook innovation gate $P_\mathrm{map}+\Sigma_k$ ties it on every chunk at
both lengths, frozen and chunk-end ($0.931/0.947$ SR at $T{=}300/450$). The gate's one residual cost is the rare confidently-wrong
capture it cannot escape. We tried three further variants, all of which
fail to improve on $P_\mathrm{map}$-only:
\begin{itemize}
  \item \textbf{Per-track gating} --- admit an observation consistent with
    \emph{any} hypothesis in the mixture, not only the MAP track. SR-neutral
    with slightly worse RMSE: the extra admissions are redundant with
    hypotheses the filter already carries.
  \item \textbf{Escape valve} --- force one ungated stride after $N$
    consecutive all-rejected updates ($N{=}3$). One chunk behind at $T{=}300$
    under the frozen readout ($0.897/0.947$ vs $0.931/0.947$ at $T{=}300/450$;
    chunk-end SR equal, one chunk swapped).
  \item \textbf{Motion-only (drift-immune) gate} --- SR-neutral, RMSE-worse.
\end{itemize}
We therefore keep the $P_\mathrm{map}$-only gate.

\paragraph{Initialization, gating, and recovery limits.}
Although consensus
chooses the absolute anchor, wrong relative shifts can still bias that
choice; the heal only replaces jumps exceeding its odometry threshold.
Birth is applied \emph{after} the gate: an accepted observation whose best
association log-evidence is below $-10$ can seed a component with log-weight
offset $-4.6$ (approximately a $0.01$ multiplier on its observation weight).
It cannot rescue a correct observation already rejected by the gate.
No explicit global relocalization is performed; consensus PSR is diagnostic,
not a reset mechanism. Recovery from an incorrect dominant mode therefore
depends on subsequent observations passing the gate.

\section{Component ablation on Bonn}
\label{app:bonn_ablation}
Table~\ref{tab:bonn_ablation} re-runs the main paper's back-end build-up and
leave-one-out on real Bonn \texttt{static\_0}, across two front-end arms. On clean
LiDAR\,\SI{270}{\degree}$+$wheel the back-end saturates as on Gibson---accumulation
alone reaches $1.000$ SR@1\,m at every $T$ and no refinement moves it. The high-drift
D455 RGB-D $+$ ORB-SLAM3 VO build-up is main-paper Table~3 (\S4.3); its
leave-one-out rows add that windowing is the
drift-bounding knob (the deployed $5/5$ beats a $15/5$ window by up to $14$\,pp,
since a shorter window accumulates less VO drift before matching).

\begin{table}[t]
\centering
\caption{\textbf{Component ablation on Bonn} (\texttt{static\_0}, last-10 SR@1\,m
$\uparrow$ / median RMSE (m) $\downarrow$; $n{=}59/29/19$ chunks at $T{=}150/300/450$;
consensus rows frozen readout, argmax and accumulation rows their single readout).
The clean LiDAR$+$wheel arm saturates (accumulation alone reaches $1.000$); the
high-drift D455$+$VO arm separates the components.}
\label{tab:bonn_ablation}
\footnotesize
\setlength{\tabcolsep}{2.5pt}
\begin{tabular}{l cc cc cc}
\toprule
 & \multicolumn{2}{c}{$T{=}150$} & \multicolumn{2}{c}{$T{=}300$} & \multicolumn{2}{c}{$T{=}450$} \\
\cmidrule(lr){2-3}\cmidrule(lr){4-5}\cmidrule(lr){6-7}
 & SR & R & SR & R & SR & R \\
\midrule
\multicolumn{7}{l}{\textit{LiDAR \SI{270}{\degree} + wheel (clean)}} \\
per-frame NCC argmax            & 0.542 & 0.20 & 0.724 & 0.12 & 0.526 & 0.21 \\
\;accumulation $\to$ full       & 1.000 & 0.08 & 1.000 & 0.09 & 1.000 & 0.10 \\
\midrule
\multicolumn{7}{l}{\textit{D455 RGB-D + VO (high-drift)}} \\
per-frame NCC argmax            & 0.153 & 7.7 & 0.069 & 7.7 & 0.105 & 10.3 \\
\;$+$ accumulation (batched)    & 0.729 & 0.21 & 0.862 & 0.20 & 0.789 & 0.34 \\
\;$+$ consensus, gate off       & 0.525 & 0.42 & 0.552 & 0.29 & 0.368 & 3.4 \\
\;$+$ GM-BF gate \& heal        & \textbf{0.780} & 0.20 & \textbf{0.966} & 0.22 & \textbf{0.842} & 0.31 \\
$-$ consensus heal              & 0.763 & 0.21 & 0.897 & 0.22 & 0.842 & 0.29 \\
$-$ $\theta$-overlap weight     & 0.780 & 0.19 & 0.931 & 0.22 & 0.895 & 0.23 \\
$W/S{=}15/5$ (wider window)     & 0.729 & 0.27 & 0.828 & 0.28 & 0.789 & 0.25 \\
\bottomrule
\end{tabular}
\end{table}

\section{Readout protocols}
\label{app:online_lag}
\begin{figure}[t]
  \centering
  \includegraphics[width=\columnwidth]{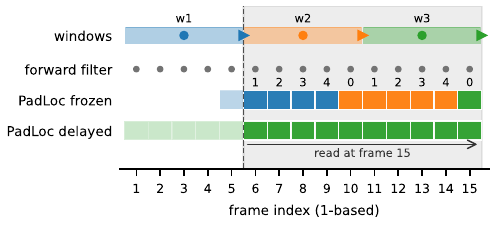}
  \caption{\textbf{Readout timing on Gibson} ($T{=}15$, $W{=}S{=}5$, 1-based
  frames). Bars: windows; dot: pose stamp (end $-2$); triangle: output
  available after the window's last frame; dashed line: first output. Grey
  band: the ten scored frames. Forward filters (F3Loc, UnLoc-\textsc{fwd}) emit
  one pose per frame. PadLoc frozen: each scored frame is coloured by the
  window that serves it, the latest window ended at or before that frame;
  numbers give the anchor age in frames. PadLoc delayed: all ten frames use the
  chunk-end anchor read at frame $T$.}
  \label{fig:timing_readout}
\end{figure}

PadLoc stamps window $w$'s pose at the window's mid-frame and emits it once
window $w{+}L$ is acquired, an acquisition delay of
$\lfloor(W{-}1)/2\rfloor{+}LS$ frames ($2$ at $W{=}S{=}5$, $L{=}0$). The
\emph{frozen} readout of the main tables scores each of the last ten frames
with the latest pose emitted at or before it, propagated by odometry and never
revised; the \emph{chunk-end} readout (PadLoc delayed, Sec.~\ref{app:delayed})
reads one anchor after frame $T$ and scores all ten frames from it ($0$--$9$
frames of look-ahead). PadLoc's first pose is emitted after frame~5, before
the first scored frame at every $T{\ge}15$ (Fig.~\ref{fig:timing_readout}).
Each readout is therefore compared only with the forward filters under the
same readout: frozen against their per-frame output (main-paper Table~1),
chunk-end against their last pose carried back
(Table~\ref{tab:delayed_tab1}). The frozen readout costs PadLoc $6$--$11$\,pp
at $T{\le}20$, $0$--$3$\,pp at $T{=}35/50$ and nothing at $T{=}100$ against
its chunk-end readout (main-paper Table~1 vs.\ Table~\ref{tab:delayed_tab1});
per cell it loses up to $31$ chunks that the chunk-end readout localizes and
gains one (\texttt{mono}, $T{=}15$), and mean RMSE over successes stays within
\SI{3}{\centi\meter}.

\paragraph{Lag sweep.}
Table~\ref{tab:online_lag} sweeps the lag $L$ on the Bonn \texttt{static\_0}
drifty-VO arm, pooling every emitted window pose of $n{=}59/29/19$ chunks
(window-pooled online SR, not the last-10 SR of the main tables). Lag mainly
sharpens the error (steady-state $p90$ \SI{0.76}{}$\to$\SI{0.53}{\meter} at
$T{=}450$ from $L{=}0$ to $\infty$); finite $L$ raises online SR by at most
$1.5$\,pp.

\begin{table}[t]
\centering
\caption{\textbf{Bounded-lag online readout on Bonn} (\texttt{static\_0} drifty-VO
arm: LiDAR\,\SI{106}{\degree} depth $+$ VO). Finite $L$ uses delayed
window outputs; $L{=}\infty$ is an offline reference. Online SR@1\,m over
emitted windows / steady-state $p90$ error (m).}
\label{tab:online_lag}
\small
\setlength{\tabcolsep}{4pt}
\begin{tabular}{r cc cc cc}
\toprule
& \multicolumn{2}{c}{$T{=}150$} & \multicolumn{2}{c}{$T{=}300$} & \multicolumn{2}{c}{$T{=}450$} \\
\cmidrule(lr){2-3}\cmidrule(lr){4-5}\cmidrule(lr){6-7}
$L$ & SR & $p90$ & SR & $p90$ & SR & $p90$ \\
\midrule
0  & 0.971 & 0.18 & 0.883 & 0.29 & 0.905 & 0.76 \\
2  & 0.974 & 0.18 & 0.884 & 0.28 & 0.911 & 0.73 \\
5  & 0.979 & 0.18 & 0.887 & 0.27 & 0.918 & 0.71 \\
10 & 0.982 & 0.18 & 0.898 & 0.26 & 0.920 & 0.68 \\
$\infty$ & 0.983 & 0.18 & 0.916 & 0.25 & 0.913 & 0.53 \\
\bottomrule
\end{tabular}
\end{table}

\section{Chunk-end readout of the main tables}
\label{app:delayed}
Main-paper Tables~1--4 report PadLoc's consensus rows under the frozen
readout. Tables~\ref{tab:delayed_tab1}--\ref{tab:delayed_tab4} give the same
runs under the chunk-end readout (PadLoc delayed, main paper \S4.1), with the
forward filters under the same readout (last forward pose carried back through
the shared transitions); argmax, accumulation, refinement, and Z-FLoc rows have
a single readout (values as in the main paper). Under this readout PadLoc
delayed leads the forward filters in 14 of 15 Gibson depth--length cells and
ties the other (F3Loc \texttt{comp\_s} at $T{=}100$). On Bonn it
leads UnLoc-\textsc{fwd} at every pooled $T$ by $+23$--$30$\,pp under wheel
odometry and $+34$--$41$\,pp with ground-truth motion (pooled GT
$0.68/0.83/0.89$); the wheel all-trial median is $0.43/0.26/\SI{0.23}{\meter}$
against UnLoc's $4.6/1.1/\SI{3.6}{\meter}$, wheel SWA $0.43/0.55/0.62$ against
$0.26/0.36/0.37$, and GT SWA $0.52/0.63/0.67$.

\begin{table}[t]
\centering
\caption{\textbf{Main-paper Table~1 under the chunk-end readout} (Gibson,
last-10 SR@1\,m $\uparrow$ / mean last-10 RMSE over successes (m) $\downarrow$).
Bold: best SR per comparison (delayed vs filter rows, accum vs batched or
refined rows), ties bold.}
\label{tab:delayed_tab1}
\scriptsize
\setlength{\tabcolsep}{1.5pt}
\begin{tabular}{l cc cc cc cc cc}
\toprule
 & \multicolumn{2}{c}{$T{=}15$} & \multicolumn{2}{c}{$T{=}20$} & \multicolumn{2}{c}{$T{=}35$} & \multicolumn{2}{c}{$T{=}50$} & \multicolumn{2}{c}{$T{=}100$} \\
\cmidrule(lr){2-3}\cmidrule(lr){4-5}\cmidrule(lr){6-7}\cmidrule(lr){8-9}\cmidrule(lr){10-11}
 & SR$\uparrow$ & R$\downarrow$ & SR$\uparrow$ & R$\downarrow$ & SR$\uparrow$ & R$\downarrow$ & SR$\uparrow$ & R$\downarrow$ & SR$\uparrow$ & R$\downarrow$ \\
\midrule
\multicolumn{11}{l}{\textit{F3Loc \texttt{mono} depth}} \\
F3Loc filter  & 0.511 & 0.25 & 0.530 & 0.24 & 0.703 & 0.19 & 0.808 & 0.18 & 0.892 & 0.18 \\
PadLoc delayed & \textbf{0.608} & 0.28 & \textbf{0.695} & 0.28 & \textbf{0.775} & 0.26 & \textbf{0.846} & 0.26 & \textbf{0.946} & 0.26 \\
\midrule
\multicolumn{11}{l}{\textit{F3Loc \texttt{comp\_s} depth}} \\
F3Loc filter  & 0.567 & 0.21 & 0.580 & 0.19 & 0.757 & 0.15 & 0.872 & 0.14 & \textbf{0.946} & 0.14 \\
PadLoc delayed & \textbf{0.675} & 0.24 & \textbf{0.770} & 0.24 & \textbf{0.910} & 0.20 & \textbf{0.962} & 0.23 & \textbf{0.946} & 0.24 \\
\midrule
\multicolumn{11}{l}{\textit{UnLoc $\mu$ depth}} \\
UnLoc fwd     & 0.810 & 0.13 & 0.865 & 0.14 & 0.919 & 0.12 & 0.949 & 0.13 & 0.973 & 0.17 \\
UnLoc refined & \textbf{0.810} & 0.17 & \textbf{0.865} & 0.17 & 0.919 & 0.17 & 0.949 & 0.17 & 0.973 & 0.20 \\
PadLoc delayed & \textbf{0.843} & 0.18 & \textbf{0.900} & 0.16 & \textbf{0.937} & 0.15 & \textbf{1.000} & 0.17 & \textbf{1.000} & 0.18 \\
PadLoc accum  & 0.795 & 0.16 & 0.860 & 0.17 & \textbf{0.946} & 0.14 & \textbf{1.000} & 0.14 & \textbf{1.000} & 0.15 \\
\bottomrule
\end{tabular}
\end{table}

\begin{table}[t]
\centering
\caption{\textbf{Main-paper Table~2 under the chunk-end readout} (Gibson
UnLoc-$\mu$ block: SWA $\uparrow$\,/\,all-trial median last-10 RMSE (m)
$\downarrow$; bold = best SWA per column).}
\label{tab:delayed_tab2}
\footnotesize
\setlength{\tabcolsep}{2.2pt}
\begin{tabular}{l ccccc}
\toprule
 & $T{=}15$ & $T{=}20$ & $T{=}35$ & $T{=}50$ & $T{=}100$ \\
\midrule
UnLoc fwd     & .703/.13 & .748/.13 & .809/.11 & .821/.12 & .809/.14 \\
UnLoc refined & .670/.18 & .717/.17 & .765/.16 & .791/.14 & .783/.17 \\
PadLoc delayed & .695/.16 & \textbf{.757}/.16 & .799/.13 & .835/.15 & .823/.18 \\
PadLoc accum  & .667/.18 & .715/.18 & .809/.14 & \textbf{.856}/.13 & .846/.14 \\
PadLoc delayed, \SI{5}{\centi\meter} & \textbf{.716}/.16 & .737/.15 & \textbf{.828}/.10 & .854/.11 & \textbf{.887}/.12 \\
\bottomrule
\end{tabular}
\end{table}

\begin{table}[t]
\centering
\caption{\textbf{Consensus rows of main-paper Table~3 and Table~\ref{tab:bonn_ablation}
under the chunk-end readout} (last-10 SR@1\,m $\uparrow$ / median last-10 RMSE (m)
$\downarrow$; Gibson UnLoc $\mu$ $n{=}268/111/37$ at $T{=}15/35/100$, Bonn
\texttt{static\_0} $n{=}59/29/19$ at $T{=}150/300/450$; full config bold).}
\label{tab:delayed_tab3}
\footnotesize
\setlength{\tabcolsep}{2.5pt}
\begin{tabular}{l cc cc cc}
\toprule
\multicolumn{7}{l}{\textit{Gibson} ($T{=}15/35/100$)} \\
\;$+$ consensus, gate/heal off  & 0.806 & 0.18 & 0.883 & 0.14 & 0.973 & 0.13 \\
\;$+$ GM-BF gate \& heal        & \textbf{0.843} & 0.16 & \textbf{0.937} & 0.13 & \textbf{1.000} & 0.18 \\
$-$ $\theta$-overlap weight     & 0.817 & 0.18 & 0.928 & 0.14 & 1.000 & 0.18 \\
$-$ $\hat b$ column drop        & 0.810 & 0.19 & 0.910 & 0.13 & 1.000 & 0.17 \\
$-$ consensus heal              & 0.832 & 0.18 & 0.910 & 0.14 & 0.973 & 0.18 \\
\midrule
\multicolumn{7}{l}{\textit{Bonn LiDAR \SI{270}{\degree} + wheel} ($T{=}150/300/450$)} \\
\;accumulation $\to$ full       & 1.000 & 0.08 & 1.000 & 0.07 & 1.000 & 0.10 \\
\midrule
\multicolumn{7}{l}{\textit{Bonn D455 RGB-D + VO} ($T{=}150/300/450$)} \\
\;$+$ consensus, gate off       & 0.576 & 0.35 & 0.655 & 0.24 & 0.421 & 2.4 \\
\;$+$ GM-BF gate \& heal        & \textbf{0.797} & 0.20 & \textbf{0.966} & 0.21 & \textbf{0.895} & 0.32 \\
$-$ consensus heal              & 0.780 & 0.21 & 0.897 & 0.21 & 0.895 & 0.33 \\
$-$ $\theta$-overlap weight     & 0.797 & 0.20 & 0.931 & 0.21 & 0.947 & 0.26 \\
$W/S{=}15/5$ (wider window)     & 0.729 & 0.27 & 0.828 & 0.25 & 0.789 & 0.26 \\
\bottomrule
\end{tabular}
\end{table}

\begin{table}[t]
\centering
\caption{\textbf{Main-paper Table~4 under the chunk-end readout} (Bonn,
monocular structural depth, cameras-only structural map; last-10 SR@1\,m;
PadLoc under wheel odometry, UnLoc-\textsc{fwd} under its ground-truth
transitions). PadLoc wins 11 of 12 cells (tie: \texttt{dynamics\_0},
$T{=}450$, $n{=}5$). $16$ rays: equal-ray control as in main-paper Table~4.
Replacing UnLoc's ground-truth transitions by wheel odometry leaves its
per-frame SR@1\,m unchanged in 20 of 22 cells and one chunk higher in the
other two (constant-$\sigma$ head, shipped and cameras-only maps).}
\label{tab:delayed_tab4}
\footnotesize
\setlength{\tabcolsep}{4pt}
\begin{tabular}{l rrr rrr}
\toprule
 & \multicolumn{3}{c}{UnLoc} & \multicolumn{3}{c}{PadLoc \textsc{cons}, wheel} \\
\cmidrule(lr){2-4}\cmidrule(lr){5-7}
scene & 150 & 300 & 450 & 150 & 300 & 450 \\
\midrule
\texttt{static\_0}      & 0.37 & 0.59 & 0.74 & \textbf{0.59} & \textbf{0.79} & \textbf{0.89} \\
\texttt{dynamics\_0}    & 0.33 & 0.43 & \textbf{0.60} & \textbf{0.67} & \textbf{0.71} & \textbf{0.60} \\
\texttt{lt\_changes\_0} & 0.28 & 0.43 & 0.21 & \textbf{0.60} & \textbf{0.71} & \textbf{0.79} \\
\texttt{lt\_ch\_dyn\_0} & 0.35 & 0.41 & 0.39 & \textbf{0.47} & \textbf{0.63} & \textbf{0.72} \\
\midrule
pooled SR@1\,m          & 0.34 & 0.48 & 0.48 & \textbf{0.56} & \textbf{0.71} & \textbf{0.79} \\
pooled, $16$ rays       &      &      &      & 0.48 & 0.67 & 0.77 \\
succ.\ med.\ RMSE (m)  & 0.21 & 0.25 & 0.21 & 0.19 & 0.20 & 0.20 \\
\bottomrule
\end{tabular}
\end{table}

\section{Real-building back-end levers}
\label{app:bonn_levers}
\paragraph{Front-end modularity table.} The main paper's \texttt{static\_0}
modularity numbers in full:

\begin{table}[t]
\centering
\caption{\textbf{Bonn \texttt{static\_0} modularity} (last-10 SR@1\,m
$\uparrow$ / median RMSE (m) $\downarrow$, frozen readout, per-chunk reset,
$n{=}29/19$). Top: deployed $\sigma$ ($1.0$ LiDAR, $0.5$ RGB-D; bold = best SR);
bottom: the same rows with $\sigma$ swapped.}
\label{tab:bonn_modular_supp}
\small
\setlength{\tabcolsep}{4pt}
\begin{tabular}{l l cc cc}
\toprule
 & & \multicolumn{2}{c}{$T{=}300$} & \multicolumn{2}{c}{$T{=}450$} \\
\cmidrule(lr){3-4}\cmidrule(lr){5-6}
depth & motion & SR & R & SR & R \\
\midrule
LiDAR \SI{270}{\degree} & VO & \textbf{1.000} & 0.10 & 0.947 & 0.12 \\
LiDAR \SI{270}{\degree} & wheel & \textbf{1.000} & 0.09 & \textbf{1.000} & 0.10 \\
LiDAR \SI{106}{\degree} & VO & 0.931 & 0.10 & 0.947 & 0.12 \\
D455 RGB-D & VO & 0.966 & 0.22 & 0.842 & 0.31 \\
\midrule
LiDAR \SI{270}{\degree}, $\sigma{=}0.5$ & VO & 0.966 & 0.09 & 0.895 & 0.14 \\
LiDAR \SI{270}{\degree}, $\sigma{=}0.5$ & wheel & 1.000 & 0.08 & 1.000 & 0.09 \\
LiDAR \SI{106}{\degree}, $\sigma{=}0.5$ & VO & 0.931 & 0.08 & 0.895 & 0.11 \\
D455 RGB-D, $\sigma{=}1.0$ & VO & 0.966 & 0.27 & 0.789 & 0.22 \\
\bottomrule
\end{tabular}
\end{table}
\paragraph{Reference-map ablation (construction and sensitivity).} We compare three
reference maps, changing one variable at a time: \textbf{(A)} the shipped
\texttt{map\_office} (unmasked one-height 2D-LiDAR; static furniture such as the
kitchenette rendered as wall); \textbf{(B)} cameras $+$ LiDAR with Mask2Former
semantic masking (wall/door/window classes kept), i.e.\ masked pixels but LiDAR
furniture retained; \textbf{(C)} cameras-only with the same masking, i.e.\ LiDAR
dropped and furniture gone. B and C are rasterized from the masked depth of the three
D455 cameras of the benchmark's \emph{mapping} run only---no test-sequence
information---on the shipped \SI{5}{\centi\meter} grid and world frame
(\texttt{facebook/mask2former-}\allowbreak\texttt{swin-large-ade-semantic}). C is \emph{not} sparser than
the shipped map (wall density $1.4\%$ vs $1.3\%$): three floor-to-ceiling cameras
cover more than the one-height scan. Under ground-truth motion with the back-end
held fixed and the structural query fixed, C beats A in 7/12 (scene,$T$) cells, ties 4,
loses 1 ($+5.4$\,pp mean SR@1\,m, frozen readout; $+7.8$\,pp under wheel odometry, 10/12
cells). The mechanism is isolated by B: masking alone changes nothing
(B${}={}$A${}={}0.400$ on \texttt{dynamics\_0} at $T{=}150$) while dropping the
LiDAR-baked furniture is the entire lever (C${}={}0.600$); gains concentrate on the
\texttt{dynamics}/\texttt{lt\_changes} sequences, where scene--map furniture mismatch
is worst. The lever is back-end-specific: the furniture-free
reference \emph{helps} PadLoc ($+5.4$\,pp) but \emph{hurts} UnLoc ($-6.3$\,pp)---the
per-frame filter cannot accumulate evidence to exploit the sparser walls-only
reference, whereas aggregate-then-match does. On the $\sigma{=}1.0$ back-end this
widens PadLoc's mean lead over UnLoc from $+18$\,pp on the shipped map to
${\sim}\!+30$\,pp; on C, PadLoc leads in 11 of 12 cells and ties the other. This study isolates the
map variable (ground-truth motion, fixed back-end, constant-$\sigma$
UnLoc); the shipped-map comparison under the published head-to-head
configuration --- wheel and ground-truth motion, UnLoc with its network
$\hat b$ --- is in the main paper and agrees in direction on every pooled
cell.

\paragraph{Borrowed $\hat b$.} Bonn's monocular structural depth carries no
uncertainty, but we can borrow UnLoc's released ViT-L head and keep only its
per-column $\hat b$ (leaving $\mu$ unchanged). On top of the $\sigma{=}0.5$ back-end
this borrowed column drop is only a small, scene-dependent gain (e.g.\ $+5$\,pp
ground-truth SR@1\,m at $T{=}150$, frozen readout) and $-3.6$ to $+1.8$\,pp elsewhere.

\section{Hilti online pose error}
\label{app:hilti_pose}
\begin{figure}[t]
  \centering
  \includegraphics[width=\columnwidth]{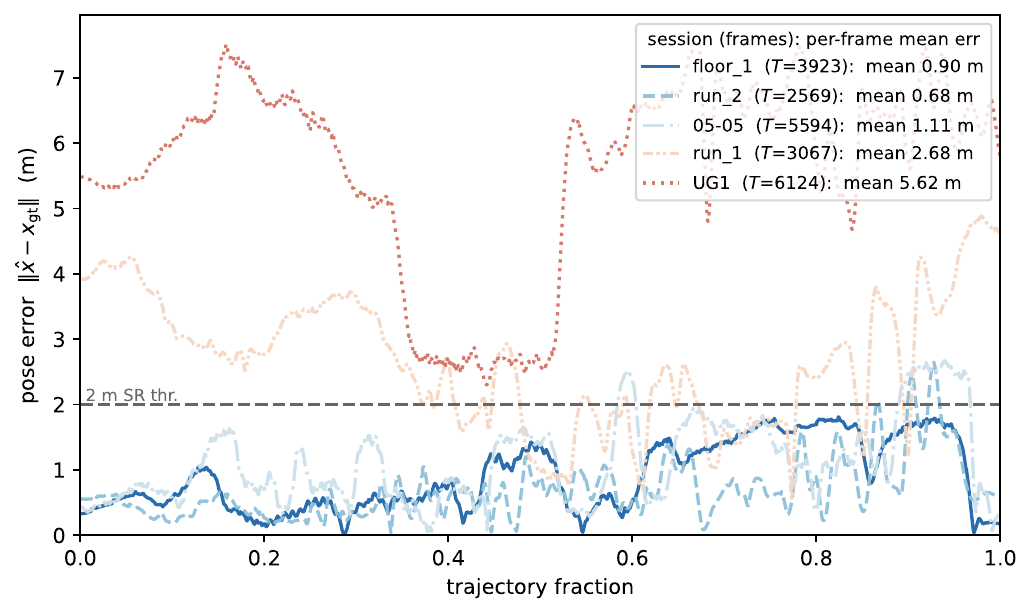}
  \caption{\textbf{Hilti online pose error} ($W/S{=}30/30$, with-windows reference).
  Per-frame error $\|\hat{x}-x_\mathrm{gt}\|$ vs.\ trajectory fraction for the five
  GT sessions; the legend lists frame count and mean error. Blue: localizing
  sessions; red: \texttt{run\_1} and \texttt{UG1}; dashed: \SI{2}{\meter} threshold.}
  \label{fig:hilti_pose}
\end{figure}

Figure~\ref{fig:hilti_pose} plots the per-frame localization error
$\|\hat{x}-x_\mathrm{gt}\|$ of the online pose stream against trajectory fraction
for all five Hilti GT sessions at the adopted $W/S{=}30/30$ (every GT frame,
$T{=}2569$--$6124$ per session---the same per-frame statistic as the mean
column of the main paper's Table~5).
The three sessions that localize (\texttt{floor\_1}, \texttt{run\_2}, \texttt{05-05})
hold a mean error at or below the \SI{2}{\meter} success threshold
(\num{0.90}, \num{0.68}, \SI{1.11}{\meter})
for essentially the whole trajectory.
\texttt{run\_1} hovers around the threshold (mean \SI{2.68}{\meter}); its
error rises above \SI{2}{\meter} where the free-space evidence fits a wrong
part of the plan as well as the true one. \texttt{UG1} stays
${\sim}\!2$--$8$\,m off for the whole run (mean \SI{5.62}{\meter}), because
its reconstructed depth collapses to amorphous blobs with nothing wall-like to
register. Section~\ref{app:hilti_fail} analyses both failures. PadLoc emits a
pose for every window, so these errors are wrong estimates, not missing
outputs.

\section{Hilti qualitative: per-window locking}
\label{app:hilti_strip}
\begin{figure}[t]
  \centering
  \includegraphics[width=\columnwidth]{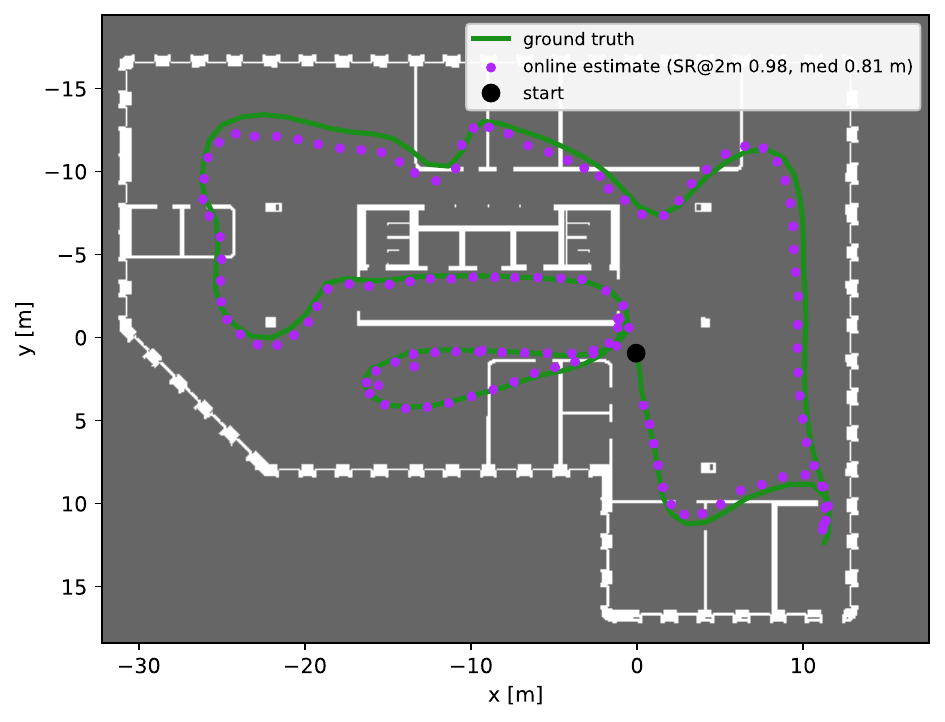}
  \caption{\textbf{Whole-run trajectory on \texttt{floor\_1}.} The online per-window
  estimate (purple) vs.\ ground truth (green) on the occupancy floor plan; the estimate
  tracks the full building loop (SR@\SI{2}{\meter} $0.98$, median \SI{0.81}{\meter}).}
  \label{fig:hilti_traj}
\end{figure}

\begin{figure*}[t]
  \centering
  \includegraphics[width=0.88\textwidth]{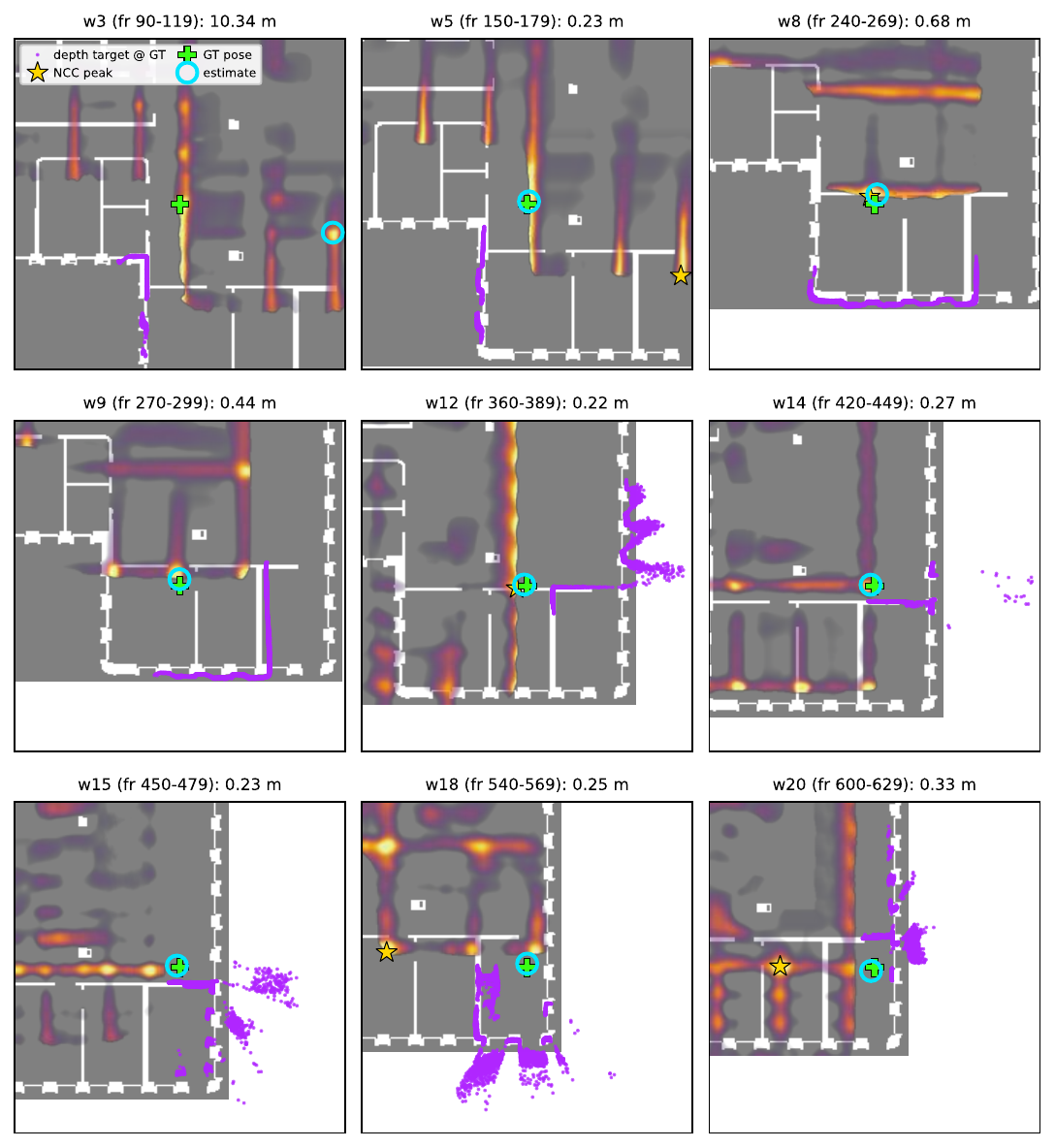}
  \caption{\textbf{Per-window locking on \texttt{floor\_1} (qualitative).} Nine
  of the 24 windows of the first 720 frames. Each panel shows the
  masked-NCC field (inferno), the depth target at the ground-truth pose (purple),
  the raw NCC peak (gold star), the decoded estimate (cyan ring), and GT (green
  plus).}
  \label{fig:hilti_strip}
\end{figure*}

Figure~\ref{fig:hilti_traj} shows the whole-run trajectory on \texttt{floor\_1},
which tracks the full building loop. Figure~\ref{fig:hilti_strip} shows a
separate run over the first \num{720} frames of the same session: nine of its
\num{24} non-overlapping \num{30}-frame windows, each cropped around the
camera, with the per-window error in the panel title. Along corridors the NCC
field often forms a ridge, because translation along a featureless corridor is
ambiguous. In the first four windows (0--3) the estimate has not yet settled
on the true corridor: in window~3, the first panel, it sits on a parallel
ridge about \SI{10}{\meter} from GT. From window~4 on, every decoded estimate
is within \SI{0.7}{\meter} of GT, five of the nine sampled ones within about
\SI{0.3}{\meter}, even where the raw NCC peak lies elsewhere (windows 5, 18
and 20).

\section{Hilti failure analysis: three regimes}
\label{app:hilti_fail}
\begin{figure*}[t]
  \centering
  \includegraphics[width=\textwidth]{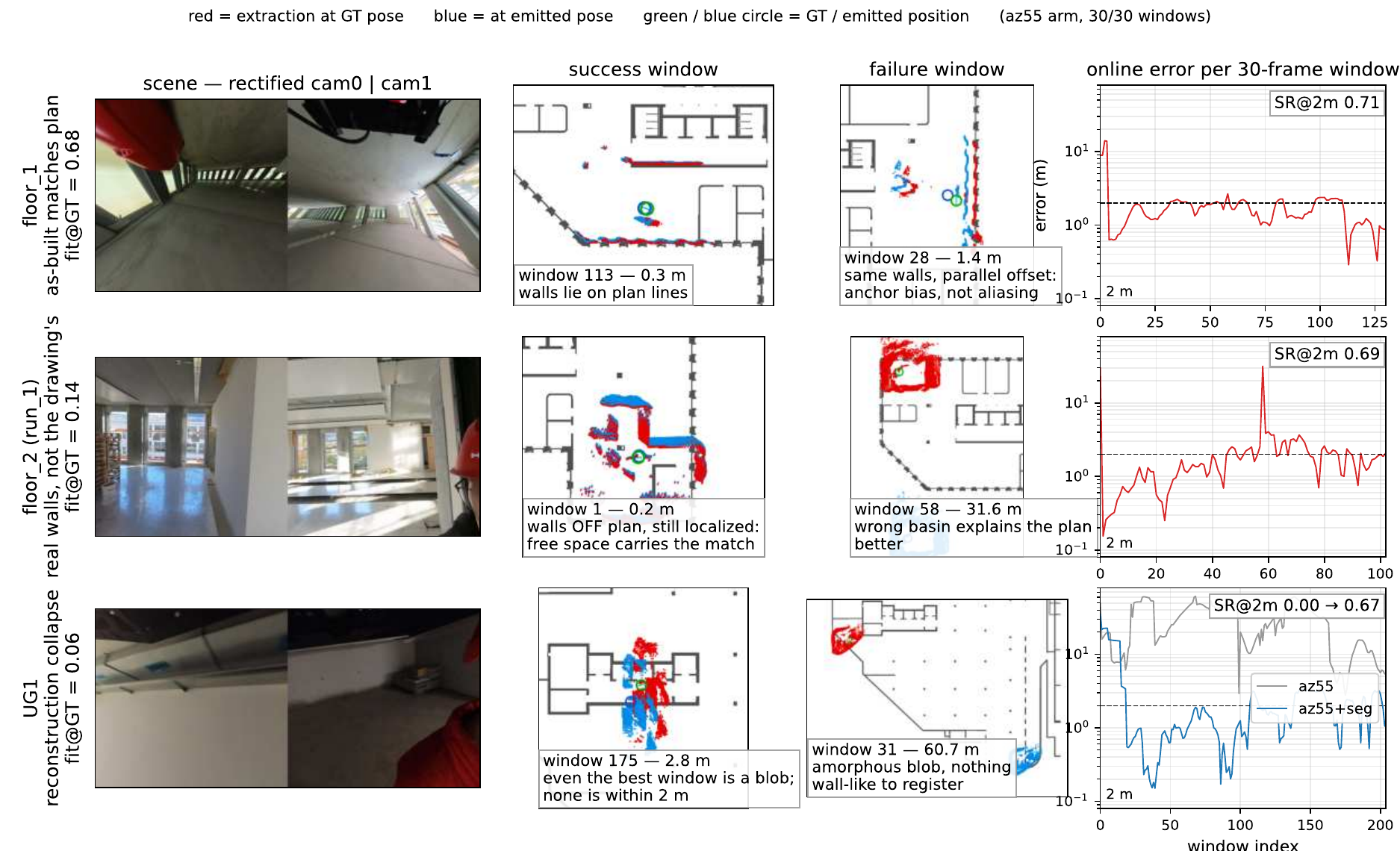}
  \caption{\textbf{Hilti failure taxonomy — three regimes.} One row per
  regime. Columns: the two rectified camera views; the wall extraction placed
  on the plan at the ground-truth (red) and emitted (blue) pose, for a success
  and a failure window; and the per-window online error ($\pm55^\circ$
  multi-crop variant; dashed: \SI{2}{\meter} success threshold).
  \texttt{floor\_1} fails by a parallel offset of a correct alignment;
  \texttt{run\_1}'s off-plan walls localize on free space and fail by
  aliasing; \texttt{UG1} reconstructs blobs, and its timeline also shows the
  recovery with learned wall segmentation ($+$seg, blue).}
  \label{fig:hilti_fail}
\end{figure*}

Placing the pipeline's own wall extraction on the plan, once at the
ground-truth pose and once at the emitted pose (Fig.~\ref{fig:hilti_fail}),
separates the five sessions into three regimes, one per physical floor. We
summarize each floor by fit@GT, the fraction of extracted wall points within
\SI{0.3}{\meter} of a plan wall when placed at ground truth.

\emph{(1) The building matches the plan} (\texttt{floor\_1}, fit@GT $0.68$).
The walls anchor localization. The remaining failures are small offsets,
parallel to the walls, of an otherwise correct alignment. They come from a
bias in the consensus anchor, not from matching a wrong place, and not from
the odometry: after alignment the VIO deviates from GT by only
\SI{0.27}{\meter} on average. Every front-end variant in main-paper Table~5
lowers accuracy here.

\emph{(2) Real walls that the drawing lacks} (the three \texttt{floor\_2}
sessions \texttt{run\_2}, \texttt{05-05} and \texttt{run\_1}, fit@GT
$0.12$--$0.17$ even at ground truth). The reconstruction is wall-like, but its
walls are drywall partitions of the ongoing fit-out, absent from the
structural drawing. Localization therefore rests on observed free space. Two
of the three sessions succeed ($0.81$--$0.95$ across front-end variants);
\texttt{run\_1} stays at $0.46$ ($0.69$ with the variants). Its failures are
aliasing: in $41$--$100\%$ of failed windows a wrong location explains the
plan at least as well as the true pose (window~58 in
Fig.~\ref{fig:hilti_fail}). Variants that widen the view help; cleaner walls
change little.

\emph{(3) Reconstruction collapse} (\texttt{UG1}, fit@GT $0.06$). Textureless
surfaces, uniform artificial light, and the operator filling a third of the
rear camera's view reduce MapAnything's output to amorphous blobs, in the best
and worst windows alike, so there is nothing wall-like to register. Wall
segmentation on the single-crop input only removes more of this sparse
target. On the $\pm55^\circ$ multi-crop input, however, it isolates the few
true structures---above all the two shaft cores, the only landmarks that tell
one part of the hall from another (removing them drops SR from $0.668$ to
$0.288$)---and recovers the floor ($0.156\to0.668$, main-paper Table~5).

\emph{Window--batch alignment.} MapAnything reconstructs non-overlapping
$30$-frame batches, and the deployed $30/30$ window uses exactly one. Windows
that mix frames from two batches, or use only five frames of one, fail on most
sessions (online SR@2\,m, \texttt{floor\_1}/\texttt{run\_2}/\texttt{05-05}/%
\texttt{run\_1}/\texttt{UG1}: $15/5$ $0.843/0.008/0.000/0.296/0.000$, $5/5$
$0.892/0.006/0.015/0.000/0.001$; a $30/5$ window reaches $0.617/0.026$ on
\texttt{floor\_1}/\texttt{run\_2}). At equal length, alignment with the
batches is what matters: a $15/15$ window, which never straddles two batches,
holds \texttt{run\_2} at $0.900$ ($0.008$ at $15/5$) and \texttt{run\_1} at
$0.985$, but loses $11$\,pp on \texttt{floor\_1} ($0.863$ vs $0.977$ at
$30/30$).

\section{Additional evidence-hygiene tools}
\label{app:hygiene}
The default configuration uses overlap-mass weighting (heading selection
only) $+$ per-column $\hat b$ drop (main paper \S3). We evaluated two further training-free tools
(frozen readout, UnLoc depth):
\begin{itemize}
  \item \textbf{Multi-view support filter} --- keep a window-cloud point only
    if $\geq\!2$ frames corroborate it within $\SI{0.2}{\meter}$ (a
    \emph{drop}, not a down-weight). Standalone (without the $\hat b$ drop)
    it changes pooled SR by $-1.8$ to $+1.1$\,pp across $T{=}15$--$100$; the
    base already reaches $1.000$ at $T{=}100$.
  \item \textbf{Reference wall unification} --- collapse aliased double-wall
    strokes to their centre line. Added to the support filter it costs
    $1.0$--$2.2$\,pp at $T{\le}50$.
\end{itemize}
On top of the default (with the $\hat b$ drop) both lower SR (support filter:
up to $2.6$\,pp at $T{=}50$; with unification: up to $8.0$\,pp at $T{=}20$; $T{=}100$
unchanged), so neither is part of the default per-frame configuration; the
Hilti batched front-end uses both.

\section{Matching resolution vs.\ depth density}
\label{app:res}
The main paper's controlled $T$-sweep holds UnLoc's $\SI{10}{\centi\meter}$
DeSDF grid so that no resolution asymmetry confounds the comparison. PadLoc's
matching cell is, however, a free knob---independent of the front-end and bounded
below only by the floor plan's own resolution. We sweep it on both datasets, down
to each map's native floor ($\SI{1}{\centi\meter}$ on Gibson,
$\SI{5}{\centi\meter}$ on Bonn), holding the other numerical settings fixed. This is a configuration
sweep: with fixed $\sigma$ in cells, changing $\delta$ also changes the
motion covariance in metres (main-paper Eq.~(5)). Differences therefore
cannot be attributed to sampling density or quantization alone.

\paragraph{Sparse depth (Gibson, UnLoc $\mu$).}
UnLoc's head emits a $40$-column fan over a $\sim\!\SI{106}{\degree}$ cone---a
\emph{sparse} target. Shrinking the cell improves precision down to $\SI{3}{\centi\meter}$ (median
last-10 RMSE falls from $\SI{0.19}{\meter}$ at $\SI{10}{\centi\meter}$ to
$\SI{0.06}{\meter}$ at $\SI{3}{\centi\meter}$, $T{=}100$, frozen readout), but success is
\emph{non-monotonic} (Table~\ref{tab:res_supp}): short-trajectory SR peaks at
$\SI{5}{\centi\meter}$ ($+1.1$\,pp at $T{=}15$), falls at $3$ and $\SI{2}{\centi\meter}$
($-5.2/-9.3$\,pp at $T{=}15$), and collapses at the
native $\SI{1}{\centi\meter}$ floor ($0.511$, $-22$\,pp). This is consistent with sparse wall sampling becoming less connected
at fine grids, but the coupled motion-noise change prevents isolating that
explanation. The shared $\SI{10}{\centi\meter}$ grid controls one important
comparison setting; it is not uniformly optimal for PadLoc.

\paragraph{Dense depth (Bonn, LiDAR).}
The $\SI{270}{\degree}$ LiDAR returns thousands of points per frame---a
\emph{dense} target. Here SR stays at $1.000$ from the native
$\SI{5}{\centi\meter}$ map floor, where the sweep reproduces the modular
\textsc{lidar+wheel} headline, up to $\SI{15}{\centi\meter}$ and drops by one
chunk only at $\SI{20}{\centi\meter}$, $T{=}450$ (Table~\ref{tab:res_bonn}).
Coarsening the dense scan costs precision rather than success: median error
rises from \SIrange{0.09}{0.10}{\meter} at \SIrange{5}{10}{\centi\meter} to
\SIrange{0.20}{0.23}{\meter} at $\SI{20}{\centi\meter}$.

\begin{table}[t]
\centering
\caption{\textbf{Sparse depth has a resolution sweet spot} (Gibson, UnLoc $\mu$
depth, 9 scenes, single frozen config, per-chunk reset). Last-10
SR@1\,m ($\uparrow$) vs.\ matching cell; per-column best in \textbf{bold}
(ties unmarked). Short-$T$ success peaks at $\SI{5}{\centi\meter}$ and falls below
it, while $T{=}20$ and $T{=}50$ favour the coarse grid.}
\label{tab:res_supp}
\small
\setlength{\tabcolsep}{5pt}
\begin{tabular}{r ccccc}
\toprule
 & \multicolumn{5}{c}{SR@1\,m $\uparrow$ at $T$} \\
\cmidrule(lr){2-6}
cell & $15$ & $20$ & $35$ & $50$ & $100$ \\
\midrule
$\SI{10}{\centi\meter}$ & 0.731 & \textbf{0.840} & 0.919 & \textbf{1.000} & 1.000 \\
$\SI{5}{\centi\meter}$  & \textbf{0.743} & 0.815 & \textbf{0.928} & 0.974 & 1.000 \\
$\SI{3}{\centi\meter}$  & 0.679 & 0.795 & 0.919 & 0.962 & 1.000 \\
$\SI{2}{\centi\meter}$  & 0.638 & 0.750 & 0.892 & 0.949 & 1.000 \\
$\SI{1}{\centi\meter}$  & 0.511 & 0.640 & 0.802 & 0.910 & 0.973 \\
\bottomrule
\end{tabular}
\end{table}

\begin{table}[t]
\centering
\caption{\textbf{Dense depth keeps its success rate on coarse cells} (Bonn,
$\SI{270}{\degree}$ LiDAR\,+\,wheel odometry, \texttt{static\_0}). Last-10
SR@1\,m ($\uparrow$) / median last-10 RMSE (m, $\downarrow$) vs.\ matching cell,
frozen readout.
SR stays at $1.000$ up to $\SI{15}{\centi\meter}$ and error grows beyond
$\SI{10}{\centi\meter}$; the native $\SI{5}{\centi\meter}$ map floor
reproduces the modular \textsc{lidar+wheel} headline; per-column best in
\textbf{bold} (ties unmarked).}
\label{tab:res_bonn}
\small
\setlength{\tabcolsep}{5pt}
\begin{tabular}{r cc cc}
\toprule
 & \multicolumn{2}{c}{$T{=}300$} & \multicolumn{2}{c}{$T{=}450$} \\
\cmidrule(lr){2-3}\cmidrule(lr){4-5}
cell & SR$\uparrow$ & RMSE$\downarrow$ & SR$\uparrow$ & RMSE$\downarrow$ \\
\midrule
$\SI{5}{\centi\meter}$  & 1.000 & \textbf{0.086} & 1.000 & 0.102 \\
$\SI{10}{\centi\meter}$ & 1.000 & 0.103 & 1.000 & \textbf{0.095} \\
$\SI{15}{\centi\meter}$ & 1.000 & 0.121 & 1.000 & 0.124 \\
$\SI{20}{\centi\meter}$ & 1.000 & 0.204 & 0.947 & 0.227 \\
\bottomrule
\end{tabular}
\end{table}

\section{Matcher latency across map and target sizes}
\label{app:runtime_sweep}
\begin{figure}[t]
  \centering
  \includegraphics[width=\columnwidth]{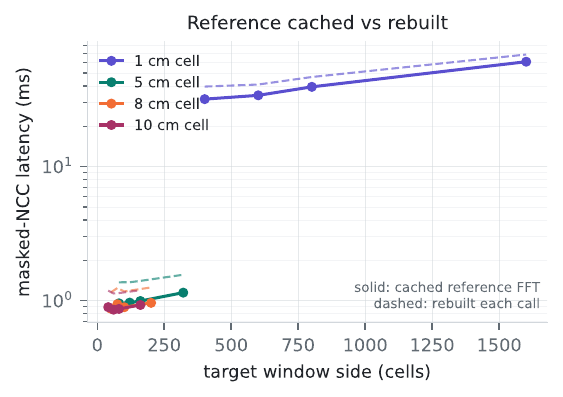}
  \caption{\textbf{Per-call masked-NCC latency} (RTX 3090, cuFFT) vs. target-window
  side on the $30\times24$\,m floor. Solid lines reuse the precomputed reference
  FFT; dashed lines rebuild it each call.}
  \label{fig:runtime_sweep}
\end{figure}

Figure~\ref{fig:runtime_sweep} sweeps the masked-NCC per-call latency on the
RTX 3090 across matching cell and target-window size, on a real
$30\times24$\,m benchmark floor (the synthetic up-scaled plans of
\S\ref{app:complexity} reach mall scale). Once the scene-constant reference FFT is cached
(solid), per-call cost varies with target size and the resulting padded FFT
dimensions at this fixed map extent, and stays sub-millisecond at the deployed $8$--$\SI{10}{\centi\meter}$ cells; only
the $\SI{1}{\centi\meter}$ grid with a large legacy target reaches tens of
milliseconds. Rebuilding the reference each call (dashed) is consistently
$\SIrange{20}{30}{\percent}$ costlier.

\section{Matcher complexity across building scales}
\label{app:complexity}
UnLoc scores every frame against its full DeSDF: $V{=}11$ rays (its depth
head's $40$ columns, downsampled to the DeSDF's \SI{10}{\degree} ray spacing,
a ${\sim}\SI{100}{\degree}$ cone within Gibson's ${\sim}\SI{106}{\degree}$
field of view) at all $O{=}36$ orientations. This Laplace-scaled $\ell_1$ scan
costs $\bigO(HWOV)$, linear in floor-plan area. PadLoc instead correlates one
fused target per $S{=}5$-frame window with the plan at $B$ candidate headings
(the plan's dominant wall orientations: $B{\approx}4$ for Manhattan plans, up
to $8$ otherwise), at $\bigO(N\log N)$ per heading.

We time both matchers on seven floor plans from $80$ to
$19{,}200$\,m$^2$ (home to mall) on one RTX 3090, taking the minimum over
repeated calls (UnLoc's released matcher; the main-paper figure). Per match,
i.e.\ one UnLoc frame or one PadLoc heading, UnLoc grows from $2.3$ to
\SI{209}{\milli\second} and PadLoc from $0.9$ to \SI{3.4}{\milli\second}:
$61\times$ at mall scale. Since PadLoc matches $B$ headings once per $S$
frames, this becomes $76\times$ per frame at $B{=}4$ ($38\times$ at $B{=}8$).

\paragraph{Where the speedup comes from.}
UnLoc's matcher loops over the $36$ orientations in Python. Replacing the loop
with one strided reduction gives bit-identical output
($\max|\Delta\mathrm{prob}|{=}0$ on all plans) and is $3.7\times$ faster at
room scale, where kernel-launch overhead dominates, but only
${\sim}1.1\times$ at mall scale, where memory traffic dominates. Against this
vectorized matcher PadLoc is still $56\times$ faster per match at mall scale.
The $\ell_1$ scan stays limited by memory bandwidth
(${\sim}\SI{18}{}\,$GFLOP/s), whereas PadLoc's FFT is limited by compute
(\SI{396}{}\,GFLOP/s). The measured gain therefore comes from constants,
implementation and PadLoc's lower matching rate ($B$ headings per $S$ frames);
it is not an asymptotic advantage of FFT matching over a scan with fixed $O$
and $V$, whose cost is linear in map area. At room scale the vectorized scan
is as fast as our FFT, which cannot drop below about \SI{0.9}{\milli\second}
because it pads to a power of two. Once the floor plan dominates the FFT size,
the target size matters little (mall: a \SI{1}{\meter} vs \SI{11}{\meter}
target costs \SI{2.6}{} vs \SI{3.4}{\milli\second}).

\paragraph{Memory and precompute.}
Values are Gibson / Bonn. Both precompute times are measured on one laptop CPU
(Apple M1~Max, 10 cores). Memory is measured where each method runs: PadLoc on
that CPU, UnLoc's released code on the RTX 3090. On the RTX 3090, PadLoc's
reference FFTs are at most $7\%$ larger and its Gibson window-update peak about
$20\%$ larger. PadLoc's back-end state consists of the cached reference FFTs
(\SIrange{1.2}{1.9}{\mega\byte} / \SI{9.9}{\mega\byte}), built from the plan
at load time in at most \SI{0.4}{\second}; \SI{13}{\byte} per cell and heading
(\SI{5}{\byte} for the heading-selection sum and its support, \SI{8}{\byte}
for the running consensus sum and count); and the \SI{560}{\byte} $K{=}10$
mixture. A window update temporarily needs another
\SIrange{5.5}{9.2}{\mega\byte} / \SI{49}{\mega\byte}. UnLoc keeps a float32
$(x,y,\theta)$ prior and posterior over $0.9$--$5.2$M / $18$M cells at its
\SI{10}{\centi\meter} grid (\SIrange{8}{44}{\mega\byte} / \SI{147}{\mega\byte}
on the GPU), plus a DeSDF that must be precomputed for each map, grid and
orientation count. Its released, unoptimized code needs \SI{4}{\minute} to
${\approx}\SI{77}{\minute}$ / ${\approx}\SI{44}{\minute}$ to build the DeSDF
(the two largest values extrapolated from every 4th / 5th grid row) and
re-uploads it every frame; with the likelihood round trip this is
\SIrange{11}{62}{\mega\byte} / \SI{288}{\mega\byte} of host--GPU traffic,
against \SIrange{1.0}{1.4}{\mega\byte} / \SI{6.6}{\mega\byte} for PadLoc's GPU
backend. Optimized code would reduce both, but the DeSDF must still be rebuilt
for each map and read in full every frame.

\paragraph{Running consensus sum.}
Every window votes on the same frame-zero anchor, so the consensus sums (main
paper, Method) never need recomputing. Shifting a field by $-o_w$, with
$o_w=g_w-g_N$ and $g_w$ the grid cell of $r_w$, equals shifting it by $-g_w$
and then by $g_N$. PadLoc therefore keeps one running sum
$\sum_w\mathrm{shift}(\tilde F_w,-g_w)$ and its count per heading: each window
adds one shifted field, a read shifts the sums by $g_N$, and a finite
\texttt{back} subtracts expired windows. One sum per heading candidate lets the
readout switch wall family without storing past fields. This is exactly equal
to re-summing every window with the same offsets ($42{,}889$ emitted window
poses on main-paper Table~1, the PadLoc cells of Table~4 and the five Hilti
$30/30$ sessions; identical chunk-end and frozen outcomes). Re-summing with the
rounded offset $o_w=\mathrm{round}((r_w-r_N)/\delta)$ instead flips at most
four trials per cell. On the UG1 grid ($1512{\times}1312$ cells), a re-summing
read costs \SIrange{8}{10}{\milli\second} per retained window
(${\ge}\SI{9.9}{\mega\byte}$ each), whereas a running-sum update and read take
\SIrange{11}{15}{\milli\second} regardless of length, with a
\SI{16}{\mega\byte} state per heading. On the Hilti UG1 $5/5$ run this cuts
wall time from $2$\,h\,$31$\,min to $33$\,min. All reported results use the
running sum.

\begin{figure}[t]
  \centering
  \includegraphics[width=\columnwidth]{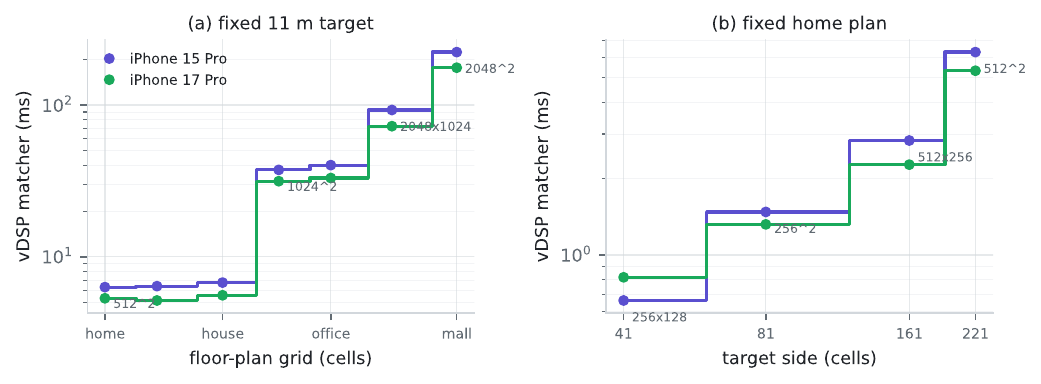}
  \caption{\textbf{On-device power-of-two staircase} (iPhone 15 Pro \& 17 Pro,
  vDSP, \SI{10}{\centi\meter}, minimum over repeated calls). \textbf{(a)}~Fixed
  \SI{11}{\meter} target, growing floor plan; \textbf{(b)}~fixed home plan,
  growing target. Both phones step through the same power-of-two FFT sizes and
  are flat within each size; the 17~Pro is ${\sim}1.2\times$ faster.}
  \label{fig:iphone}
\end{figure}

\paragraph{On-device confirmation (iPhone 15 Pro and 17 Pro).}
We ran the same masked-NCC kernel on an iPhone 15 Pro and an iPhone 17 Pro
(Accelerate/vDSP; Fig.~\ref{fig:iphone}, minimum over repeated calls). Its
cost rises in steps, one per power-of-two FFT size. With our \SI{11}{\meter}
target fixed and growing floor plans, the FFT steps through $512^2$, $1024^2$
and $2048^2$ (15~Pro \SI{6.3}{}/\SI{38}{}/\SI{224}{\milli\second}; 17~Pro
\SI{5.3}{}/\SI{32}{}/\SI{176}{\milli\second}); with the home plan fixed and
growing targets, it steps from $256^2$ to $512^2$. Within one FFT size the cost
is flat: on each phone three floor plans all cost the same at $512^2$
(${\sim}\SI{6.3}{}$ / \SI{5.3}{\milli\second}), and on the mall plan every
target we use ($41$--$221$ cells) falls into the $2048^2$ size, so the generous
target window that the narrow field of view of a single depth frame motivates
costs almost nothing. Precomputing the reference FFTs is
\SIrange{25}{40}{\percent} faster than computing them per call, and the newer
phone is uniformly ${\sim}1.2\times$ faster. At deployed room-to-floor scale
($512^2$--$1024^2$) the matcher takes \SIrange{5}{40}{\milli\second} per
heading, at mall scale \SIrange{176}{224}{\milli\second}. These numbers cover
the matching kernel only, not rasterization, heading estimation, temporal
decoding or depth inference, so they do not measure full-system throughput on
the phone.

\section{Unconditional metrics: full grid}
\label{app:swa_grid}
Table~\ref{tab:swa_grid} extends main-paper Table~2 (SWA and all-trial
median) to the two F3Loc depth sources and to the individual Bonn sequences.
Two patterns recur.
First, F3Loc's filter succeeds less often but is more precise when it
succeeds. It trails or ties PadLoc in SR in all ten Gibson cells, yet its
successful trials converge tightly, so at $T{=}100$ its SWA exceeds PadLoc's
at \SI{10}{\centi\meter} on both depth sources. PadLoc's
\SI{5}{\centi\meter} configuration removes the same pattern on the UnLoc-depth
block (main-paper Table~2).
Second, the all-trial median shows what success-conditioned RMSE hides: the
errors are bimodal (tens of centimetres for successes, metres for failures),
so once fewer than half the trials succeed, the median jumps to metres,
e.g.\ \texttt{lt\_ch\_dyn} with wheel odometry at $T{=}150$ (SR $0.45$,
median \SI{1.66}{\meter}).
Two main-paper rows have no entry here: Z-FLoc publishes no per-trial outputs
(its row is quoted from its paper), and PadLoc's MapAnything-depth row
combines two run archives. The per-archive numbers are in the released
experiment record.

\begin{table}[t]
\centering
\caption{\textbf{Unconditional-metric grid}: SWA $\uparrow$\,/\,all-trial
median last-10 RMSE (m) $\downarrow$ for the runs behind the main paper's SR
cells. Top: Gibson, F3Loc depth sources (filter: per-frame readout; PadLoc:
frozen). Bottom: Bonn per sequence, PadLoc \textsc{cons} with wheel and GT
motion (frozen) vs UnLoc (own $\hat b$, per-frame forward readout).}
\label{tab:swa_grid}
\scriptsize
\setlength{\tabcolsep}{1.6pt}
\begin{tabular}{l ccccc}
\toprule
\multicolumn{6}{l}{\textit{Gibson, F3Loc depth}} \\
 & $T{=}15$ & $T{=}20$ & $T{=}35$ & $T{=}50$ & $T{=}100$ \\
\midrule
F3Loc filter (\texttt{mono})  & .212/2.32 & .264/1.89 & .435/0.41 & .577/0.22 & .730/0.17 \\
PadLoc frozen (\texttt{mono}) & .352/0.85 & .423/0.47 & .562/0.29 & .620/0.26 & .685/0.25 \\
F3Loc filter (\texttt{comp\_s})  & .289/1.95 & .337/1.34 & .524/0.25 & .687/0.13 & .804/0.13 \\
PadLoc frozen (\texttt{comp\_s}) & .420/0.54 & .513/0.32 & .687/0.19 & .736/0.19 & .737/0.19 \\
\midrule
\multicolumn{6}{l}{\textit{Bonn per-sequence} (columns $T{=}150/300/450$)} \\
 & \multicolumn{1}{c}{\texttt{static}} & \multicolumn{1}{c}{\texttt{dyn}} & \multicolumn{1}{c}{\texttt{lt\_ch}} & \multicolumn{1}{c}{\texttt{lt\_ch\_dyn}} & pooled \\
\midrule
UnLoc $T{=}150$   & .260/5.33 & .260/4.39 & .164/7.29 & .243/5.27 & .231/5.30 \\
UnLoc $T{=}300$   & .448/0.31 & .332/6.89 & .322/3.62 & .305/3.80 & .361/1.03 \\
UnLoc $T{=}450$   & .528/0.28 & .406/0.40 & .155/3.96 & .303/7.23 & .352/3.50 \\
wheel $T{=}150$   & .437/0.44 & .471/0.46 & .490/0.30 & .356/1.66 & .427/0.46 \\
wheel $T{=}300$   & .596/0.23 & .514/0.42 & .574/0.21 & .482/0.26 & .547/0.25 \\
wheel $T{=}450$   & .693/0.23 & .479/0.24 & .621/0.23 & .576/0.28 & .618/0.24 \\
GT $T{=}150$      & .519/0.33 & .521/0.35 & .516/0.29 & .430/0.51 & .490/0.34 \\
GT $T{=}300$      & .631/0.28 & .582/0.40 & .612/0.23 & .661/0.24 & .632/0.26 \\
GT $T{=}450$      & .779/0.18 & .781/0.21 & .561/0.26 & .609/0.28 & .670/0.24 \\
\bottomrule
\end{tabular}
\end{table}

\section{Heading error, wall-family failures, and per-scene intervals}
\label{app:heading_stats}
Table~\ref{tab:heading_stats} reports the chunk-anchor heading error of the
Gibson cells under the frozen readout (per-chunk outputs of the exact-NCC runs,
SR-exact with main-paper Table~1). The error is wrapped to
$[0^\circ,180^\circ]$; a chunk is \emph{wrong-family} when it is at least
$45^\circ$, i.e.\ the directed wall-orientation candidate is off by a
quarter turn or more. Three facts. (i)~With the right family the heading is
tight: on UnLoc depth the median is $1.3$--$2.1^\circ$ at every $T$, at or
below the $5^\circ$ pre-pass bin, with p90 under $6^\circ$. (ii)~Failures
are discrete family flips, not drift: the p90 jumps to $90$--$173^\circ$ at
$T{=}15$, and a wrong family essentially never coincides with a position
success ($0.0$--$0.5\%$ of chunks). (iii)~Wrong-family chunks account for
$37/72$ ($51\%$) of the $T{=}15$ failures and $19/32$ ($59\%$) at $T{=}20$
on UnLoc depth, and for $77/135$ on F3Loc \texttt{mono}---the directed
heading ambiguity named in the main-paper Limits, quantified; the remaining
failures are right-family chunks whose serving window is off at some of the
last ten frames, mostly early-window misses that the chunk-end readout
recovers ($30$ of $35$ at $T{=}15$ on
UnLoc depth, main-paper Table~1 vs.\ Table~\ref{tab:delayed_tab1}). By $T{=}50$
no wrong-family chunk remains on UnLoc depth; the two $T{=}100$ failures on
each F3Loc arm are wrong-family.

Table~\ref{tab:per_scene} gives the per-scene SR of the UnLoc-depth frozen
cells with chunk counts, and the pooled SR with Wilson $95\%$ intervals.
Per-scene SR spans $0.62$--$0.95$ at $T{=}15$ (Stokes lowest,
Sumas highest) and is $1.00$ in every scene at $T{=}100$. The pooled
intervals, $[0.68,0.78]$ at $T{=}15$ and $[0.78,0.88]$ at $T{=}20$, overlap
UnLoc-\textsc{fwd}'s per-frame interval at $T{=}15$ ($[0.60,0.71]$, Wilson at
the shared $n$), are disjoint from it at $T{=}20$ ($[0.64,0.77]$), and overlap
again from $T{=}35$ on. These intervals only describe the spread: they treat chunks as
independent, although chunks from the same scene are correlated, and
they are not a paired test between methods. Single Bonn sequences are
even less certain. At $T{=}450$, \texttt{dynamics\_0} has only $n{=}5$
chunks, and its $3/5$ successes give an interval of $[0.23,0.88]$. This
is why the main paper reports Bonn pooled over the four sequences and
gives the chunk counts.

\begin{table}[t]
\centering
\caption{\textbf{Heading error and wall-family failures} (Gibson cells,
frozen readout; chunk-anchor heading vs.\ ground truth, wrapped to
$[0^\circ,180^\circ]$). Wrong family: error $\ge45^\circ$. Last column:
failed chunks that are wrong-family / all failed chunks.}
\label{tab:heading_stats}
\scriptsize
\setlength{\tabcolsep}{2.2pt}
\begin{tabular}{l r r c c c c}
\toprule
depth & $T$ & $n$ & SR@1\,m & med/p90 ($^\circ$) & wrong fam. & fail.\ wrong-fam. \\
\midrule
UnLoc $\mu$ & 15 & 268 & 0.731 & 2.6 / 90.1 & 13.8\,\% & 37/72 \\
UnLoc $\mu$ & 20 & 200 & 0.840 & 1.7 / 13.1 & 9.5\,\% & 19/32 \\
UnLoc $\mu$ & 35 & 111 & 0.919 & 1.7 / 7.3 & 6.3\,\% & 7/9 \\
UnLoc $\mu$ & 50 & 78 & 1.000 & 1.3 / 3.5 & 0.0\,\% & -- \\
UnLoc $\mu$ & 100 & 37 & 1.000 & 1.4 / 3.6 & 0.0\,\% & -- \\
\midrule
\texttt{comp\_s} & 15 & 268 & 0.563 & 3.4 / 172.7 & 25.7\,\% & 68/117 \\
\texttt{comp\_s} & 20 & 200 & 0.685 & 3.0 / 94.2 & 18.5\,\% & 36/63 \\
\texttt{comp\_s} & 35 & 111 & 0.883 & 1.9 / 10.9 & 8.1\,\% & 9/13 \\
\texttt{comp\_s} & 50 & 78 & 0.949 & 1.5 / 10.2 & 3.8\,\% & 3/4 \\
\texttt{comp\_s} & 100 & 37 & 0.946 & 1.5 / 5.6 & 5.4\,\% & 2/2 \\
\midrule
\texttt{mono} & 15 & 268 & 0.496 & 5.4 / 169.7 & 29.1\,\% & 77/135 \\
\texttt{mono} & 20 & 200 & 0.585 & 3.9 / 154.9 & 26.0\,\% & 51/83 \\
\texttt{mono} & 35 & 111 & 0.766 & 2.7 / 94.2 & 18.0\,\% & 20/26 \\
\texttt{mono} & 50 & 78 & 0.821 & 2.4 / 86.0 & 12.8\,\% & 10/14 \\
\texttt{mono} & 100 & 37 & 0.946 & 1.5 / 5.1 & 5.4\,\% & 2/2 \\
\bottomrule
\end{tabular}
\end{table}

\begin{table}[t]
\centering
\caption{\textbf{Per-scene SR@1\,m and pooled Wilson 95\,\% intervals}
(UnLoc $\mu$ depth, PadLoc frozen; chunk counts in parentheses).}
\label{tab:per_scene}
\scriptsize
\setlength{\tabcolsep}{2.5pt}
\begin{tabular}{l c c c c c}
\toprule
scene & $T{=}15$ & $T{=}20$ & $T{=}35$ & $T{=}50$ & $T{=}100$ \\
\midrule
Spencerville & 0.72 (39) & 0.79 (29) & 0.81 (16) & 1.00 (11) & 1.00 (5) \\
Spotswood & 0.75 (20) & 0.73 (15) & 1.00 (8) & 1.00 (6) & 1.00 (3) \\
Springhill & 0.74 (27) & 0.90 (20) & 0.91 (11) & 1.00 (8) & 1.00 (4) \\
Stilwell & 0.71 (45) & 0.85 (34) & 0.89 (19) & 1.00 (13) & 1.00 (6) \\
Stokes & 0.62 (24) & 0.78 (18) & 0.70 (10) & 1.00 (7) & 1.00 (3) \\
Sumas & 0.95 (20) & 0.93 (15) & 1.00 (8) & 1.00 (6) & 1.00 (3) \\
Superior & 0.67 (43) & 0.84 (32) & 1.00 (18) & 1.00 (12) & 1.00 (6) \\
Swormville & 0.82 (33) & 0.88 (25) & 1.00 (14) & 1.00 (10) & 1.00 (5) \\
Woonsocket & 0.65 (17) & 0.83 (12) & 1.00 (7) & 1.00 (5) & 1.00 (2) \\
\midrule
pooled & 0.731 (268) & 0.840 (200) & 0.919 (111) & 1.000 (78) & 1.000 (37) \\
Wilson 95\,\% & [0.68, 0.78] & [0.78, 0.88] & [0.85, 0.96] & [0.95, 1.00] & [0.91, 1.00] \\
\bottomrule
\end{tabular}
\end{table}

\end{document}